%% file: main.tex
\documentclass[10pt,twocolumn,letterpaper]{article}

\usepackage{cvpr}              % To produce the CAMERA-READY version
\input{preamble}

\definecolor{cvprblue}{rgb}{0.21,0.49,0.74}
\usepackage[pagebackref,breaklinks,colorlinks,citecolor=cvprblue]{hyperref}
\usepackage{float}
\usepackage{ulem}

\usepackage[accsupp]{axessibility}

\def\paperID{1338} % *** Enter the Paper ID here
\def\confName{CVPR}
\def\confYear{2024}

\title{WaveFace: Authentic Face Restoration with Efficient Frequency Recovery}

\author{Yunqi Miao\\
University of Warwick\\
{\tt\small Yunqi.Miao.1@warwick.ac.uk}
\and
Jiankang Deng\\
Imperial College London\\
{\tt\small jiankangdeng@gmail.com}
\and
Jungong Han\\
University of Sheffield\\
{\tt\small jungonghan77@gmail.com}
}

\begin{document}
% \maketitle
\twocolumn[{%
\renewcommand\twocolumn[1][]{#1}%
\maketitle

\renewcommand{\tabcolsep}{.5pt}
\begin{figure}[H]
\hsize=\textwidth
\vspace{-0.35in}
\begin{minipage}{\textwidth}

\begin{center}
\begin{tabular}{ccccccc}
   \vspace{-1.0mm}
   \includegraphics[scale=0.48]{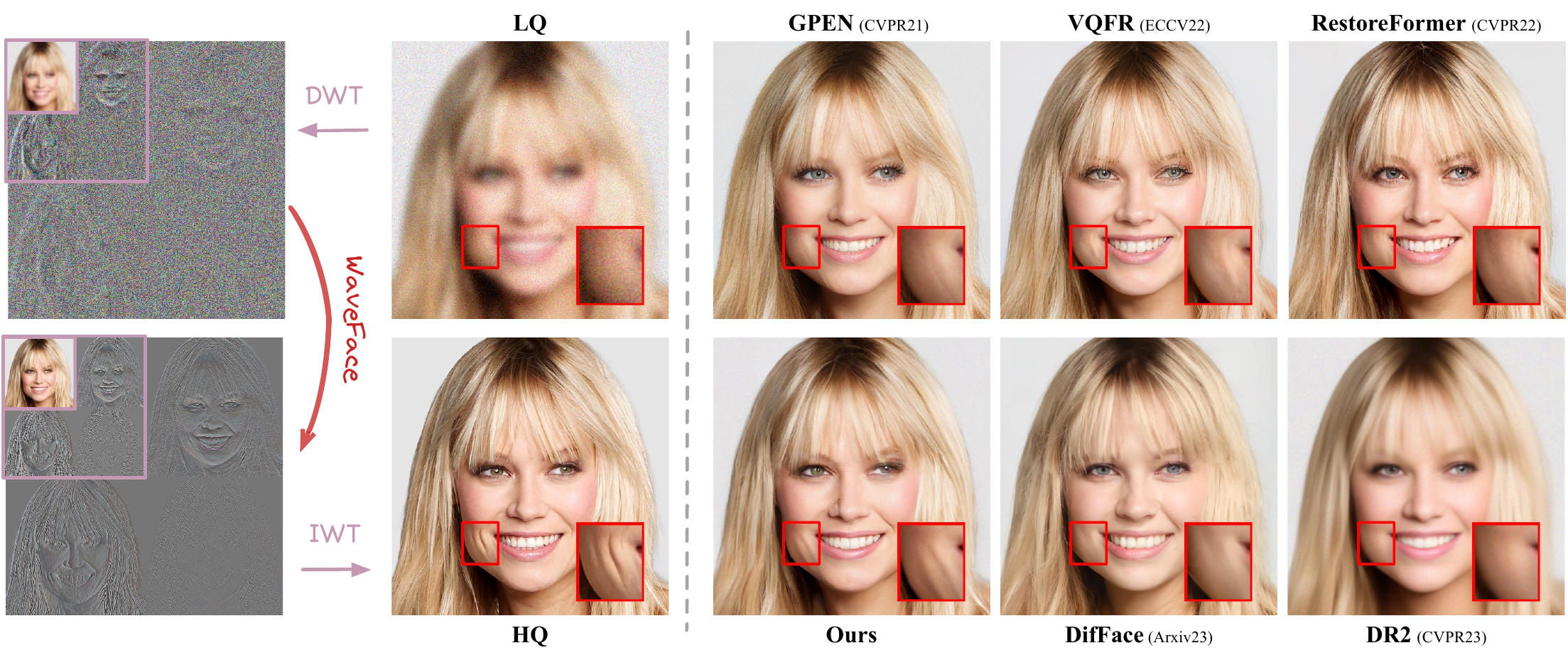}
\end{tabular}
\end{center}
\vspace{-6mm}
\caption{Left: Illustration of our frequency-aware BFR scheme. Restoration is performed in the frequency domain instead of the pixel domain. Right: Comparisons with state-of-the-art face restoration methods on degraded images. Previous methods struggle to restore facial details or the original identity while our \ours achieves a good balance of realness and fidelity with fewer artifacts.}
\label{fig:intro}
\end{minipage}
\end{figure}%
}]

\input{sec/0_abstract} 
\input{sec/1_intro}

\input{sec/2_related}

\input{sec/3_method}

\input{sec/4_exper}

\input{sec/5_con}
\input{sec/X_suppl}

% \clearpage
{
    \small
    \bibliographystyle{ieeenat_fullname}
    \bibliography{main}
}

\end{document}

%% file: preamble.tex
\usepackage[dvipsnames]{xcolor}
\usepackage{colortbl}

\newcommand{\ours}[0]{{\color{black}WaveFace}\xspace}

\usepackage{times}
\usepackage{graphicx}
\usepackage{amsmath}
\usepackage{amssymb}

\usepackage{subcaption}
\usepackage{multirow}
\usepackage{bm}
\usepackage{ifthen}

\usepackage{makecell}
\usepackage{cite}
\usepackage{mathtools}

\usepackage{array}
\newcolumntype{C}[1]{>{\centering\arraybackslash}p{#1}}

\newcommand{\bgcblue}[1]{{\cellcolor[HTML]{EDF6FF}{#1}}} % background color
\newcommand{\bgcgrey}[1]{{\cellcolor[HTML]{FAF0E6}{#1}}} % background color
\newcommand{\bred}[1]{\textbf{\color[HTML]{CB0000}{#1}}}
\newcommand{\arrowpur}[1]{\color[HTML]{008B8B}\scriptsize{$\uparrow$}{#1}}

\DeclareMathOperator{\N}{\mathcal{N}}
\DeclareMathOperator{\I}{\mathbf{I}}

%% file: sec/0_abstract.tex
\begin{abstract}
Although diffusion models are rising as a powerful solution for blind face restoration, they are criticized for two problems:
1) slow training and inference speed, and 2) failure in preserving identity and recovering fine-grained facial details.
In this work, we propose \ours to solve the problems in the frequency domain, where low- and high-frequency components decomposed by wavelet transformation are considered individually to maximize authenticity as well as efficiency.
The diffusion model is applied to recover the low-frequency component only, which presents general information of the original image but 1/16 in size.
To preserve the original identity, the generation is conditioned on the low-frequency component of low-quality images at each denoising step.
Meanwhile, high-frequency components at multiple decomposition levels are handled by a unified network, which recovers complex facial details in a single step.
Evaluations on four benchmark datasets show that:
1) \ours outperforms state-of-the-art methods in authenticity, especially in terms of identity preservation, and
2) authentic images are restored with the efficiency 10$\times$ faster than existing diffusion model-based BFR methods.
\end{abstract}

%% file: sec/1_intro.tex
\section{Introduction}
\label{sec:intro}
Blind face restoration (BFR) aims to recover high-quality (HQ) facial images from various degradations, including down-sampling, blurriness, noise, and compression artifact~\cite{bulat2018learn,zhang2022edface,wang2022survey}.
BFR is challenging since the types of degradation are generally unknown in real-world scenarios.
Improvements in the restoration quality over the past few years mainly benefit from the usage of multiple facial priors.
Geometric priors, including facial landmarks~\cite{chen2018fsrnet} and parsing maps~\cite{shen2018deep, chen2021progressive}, are employed to provide explicit facial structure information.
Reference priors, such as HQ images/features~\cite{zhou2022codeformer,wang2022restoreformer,gu2022vqfr}, are used as guidance during facial restoration.
Generative priors are generally obtained by pre-trained generation models such as StyleGAN~\cite{karras2019stylegan}, which facilitate the recovery of realistic textures~\cite{chen2021progressive, wang2021gfpgan, yang2021gpen}.
Inspired by the superior generative ability of diffusion models (DMs)~\cite{ho2020denoising, song2020denoising}, how to unleash its potential in authentic face restoration has gained much attention.

Despite plausible restoration results being achieved, existing DMs based methods~\cite{yue2022difface, wang2023dr2} generally suffer from two problems:
1) DMs are trained on large images (512$\times$512), which requires massive computational resources for both training and inference, as thousands of iterations are required to obtain adequate outputs.
DifFace~\cite{yue2022difface}, for example, takes about 25 seconds to sample an image from noise.
2) Low-quality (LQ) images are firstly \textbf{diffused} to an intermediate step and then \textbf{denoised} by an unconditional diffusion model to obtain HQ counterparts~\cite{yue2022difface}.
However, as shown in \cref{fig:intro}, such an unconditional generation will lead to great uncertainty in restored images, which means the original identity and facial details such as wrinkles cannot be well preserved.

To solve the above problems, we transfer the BFR task from the pixel domain to the frequency domain via Discrete Wavelet Transform (DWT).
As shown in \cref{fig:intro}, an image can be decomposed and reconstructed by its four quarter-sized sub-bands: low- and high-frequency components without sacrificing information.
Low-frequency sub-band mainly contains general information such as face structure while high-frequency ones contain rich facial details.
By exponentially shrinking the size of input images, restoration within the frequency domain not only speeds up both training and inference but also maintains the authenticity of restoration.
In the paper, we devise an efficient BFR method \ours that restores authentic HQ images by recovering its frequency components.

\ours consists of a Low-frequency Conditional Denoising (LCD) module and a High-Frequency Recovery (HFR) module.
A diffusion model is used in LCD to restore the low-frequency sub-band of HQ images whose size is only 1/16 of the original image.
In addition to smaller inputs, LCD leverages LQ counterparts as a condition throughout the generation to preserve the original identity.
Meanwhile, HFR recovers high-frequency sub-bands decomposed at multiple DWT levels simultaneously within one step.
With the frequency components restored by two modules, authentic images can be reconstructed via discrete inverse wavelet transform (IWT) within 1 ms.
The contributions can be summarized as follows:
\begin{itemize}[topsep=0pt,parsep=0pt,leftmargin=18pt]
    \item We propose an efficient blind face restoration approach, \ours, that restores authentic images by recovering their frequency components individually.
    \item A conditional diffusion model is adopted to restore the low-frequency component, which is 1/16 the size of the original image. 
    \item A one-pass network is used to recover high-frequency sub-bands decomposed at multiple DWT levels simultaneously.
    \item Comprehensive experiments demonstrate the superiority of methods in both efficiency and authenticity.
\end{itemize}

%% file: sec/2_related.tex
\section{Related Work}
\label{sec:related}
\subsection{Diffusion models}
Diffusion Models (DMs) are emerging generative models that corrupt the data with the successive addition of Gaussian noise during the diffusion process and then learn to recover the data during denoising.
State-of-the-art DMs~\cite{ho2020denoising} have revealed the potential in CV tasks, such as image deblurring~\cite{kawar2022ddrm,chung2022dps} and image super-resolution~\cite{saharia2022sr3,rombach2022ldm}.

\noindent\textbf{Accelerating DMs.}
Despite being powerful in generation, DDPM~\cite{ho2020denoising} has the downside of low inference speed, which requires thousands of steps for sampling.
To accelerate inference, DDIM~\cite{nichol2021improved} adopts a non-Markovian diffusion process that allows step skipping during sampling.

\noindent\textbf{Conditional DMs.}
Unconditional DM-based generation leads to great uncertainty in outputs, which means they generally fail to preserve the original identity and fine-grained facial details.
Therefore, conditions are injected by cross-attention layer~\cite{rombach2022ldm}, adaptive normalization layer~\cite{preechakul2022diffae}, or concatenation operation~\cite{saharia2022sr3} to control the characteristic of the synthesized images.

\subsection{Blind face restoration}
Facial priors exploited in blind face restoration (BFR) can be categorized into three types: geometric priors, reference priors, and generative priors.

\noindent{\bf Geometric priors} based methods leverage knowledge from facial landmark~\cite{chen2018fsrnet,deng2019menpo}, facial parsing maps~\cite{shen2018deep, chen2021progressive,zheng2022decoupled}, and facial component heatmaps~\cite{yu2018face}.
However, they tend to show inferior performance since degraded images fail to provide accurate and adequate structural information.

\noindent{\bf Reference priors} based methods either leverage a high-quality (HQ) reference image sharing the same identity as the degraded one~\cite{li2018learning} or a pre-constructed dictionary storing HQ facial features~\cite{wang2022restoreformer, gu2022vqfr, zhou2022codeformer}.
Firstly, a vector-quantized (VQ) codebook is pre-trained on HQ faces via VQ-GAN~\cite{esser2021vqgan} to provide rich facial details.
Features of degraded inputs are then fused with the prior either at image~\cite{wang2022restoreformer} or latent~\cite{gu2022vqfr,zhou2022codeformer} level for the generation of HQ counterparts.
However, reference prior is restricted by the size of the codebook, which limits the diversity and richness of generated images.

\noindent{\bf Generative priors} encapsulated in pretrained face models~\cite{karras2019stylegan, karras2020stylegan2} are also leveraged to restore faithful faces~\cite{chen2021progressive,yang2021gpen,wang2021gfpgan}. 
PSFRGAN~\cite{chen2021progressive} modulates features at different scales progressively with facial parsing maps to achieve semantic-aware style transformation.
GFP-GAN~\cite{wang2021gfpgan} and GPEN~\cite{yang2021gpen} adopt the pre-trained StyleGAN as a decoder and achieve a good balance between visual quality and fidelity of restored images.
Recently, diffusion models~\cite{ho2020denoising,song2020denoising} have shown the powerful generative ability in restoring HQ content from noisy images.
Inspired by the ability, DifFace~\cite{yue2022difface} firstly maps low-quality (LQ) inputs into an intermediate step of the denoising process, from which its HQ counterpart is recursively sampled.
DR2~\cite{wang2023dr2} employs the diffusion model to remove degradations, followed by a super-resolution model to obtain HQ counterparts.

Nevertheless, previous DM-based BFR methods are performed in the pixel domain, where large inputs require massive computational resources for both training and inference. 
Besides, they both apply the unconditional scheme that leads to significant uncertainty in restored results.

%% file: sec/3_method.tex
%-------------------------------
\begin{figure*}
    \centering
    \includegraphics[width=0.98\textwidth]{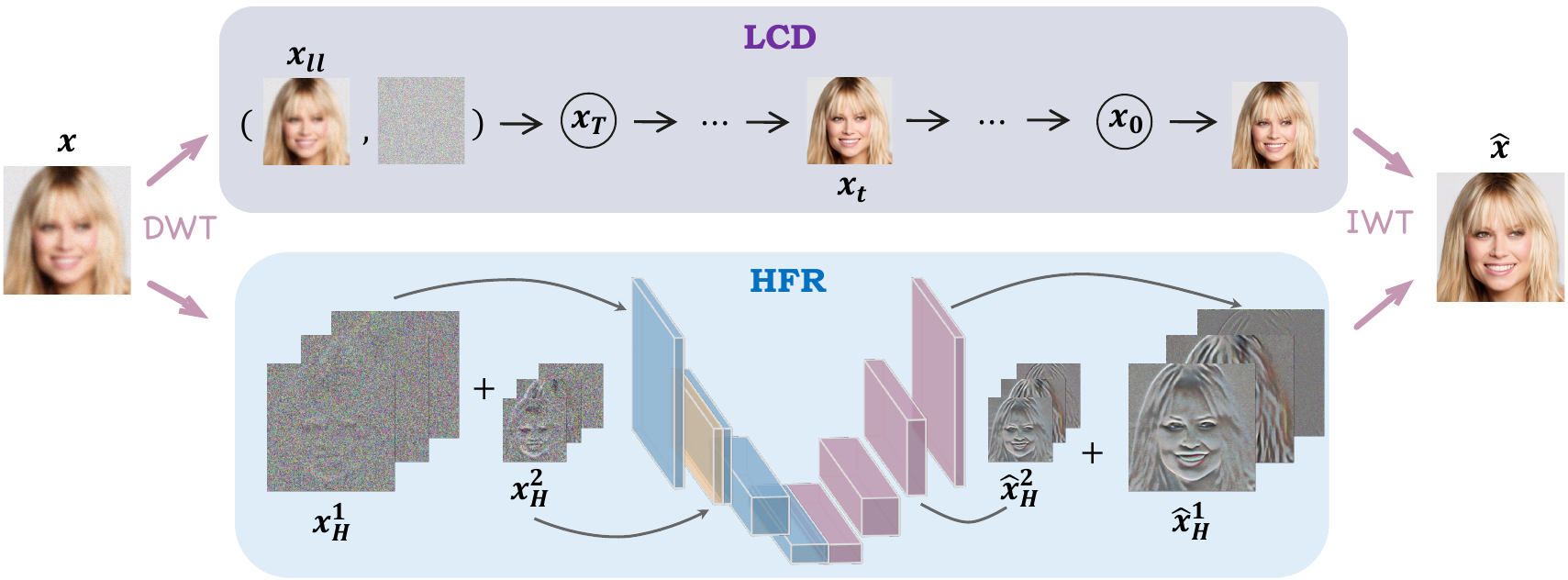}
    \caption{Overall framework of \ours. It consists of a Low-frequency Conditional Denoising (LCD) module and a High-Frequency Refinement (HFR) module. {LCD (\cref{sec:LCD}) predicts clean samples $\bm{x}_{0}$ from noise conditioned on LQ inputs through $T$ steps.} Meanwhile, high-frequency sub-bands are concatenated as HFR (\cref{sec:HFR}) inputs to recover vivid facial details.
    {The predicted frequency components are projected back to the image via IWT.}}
    \label{fig:framework}
    \vspace{-3mm}
\end{figure*}
%-------------------------------
\section{Preliminary}
\noindent \textbf{Discrete Wavelet Transformation (DWT).}\label{sec:dwt}
DWT decomposes an image into low-frequency and high-frequency sub-bands.
The low-frequency component presents general information while high-frequency ones express facial details in the vertical, horizontal, and diagonal directions.
In the paper, we use Haar wavelet for DWT, which is widely used in real-world applications due to simplicity~\cite{phung2023wavediff}.

Given an image $\bm{x} \in \mathbb{R}^{H \times W \times 3}$, its low-frequency sub-band $\bm{x}_{ll}^1 \in \mathbb{R}^{H/2 \times W/2 \times 3}$ and high-frequency sub-bands $\bm{x}_{lh}^1, \bm{x}_{hl}^1$, and $\bm{x}_{hh}^1 \in \mathbb{R}^{H/2 \times W/2 \times 3}$ can be decomposed by:
\begin{equation}
    \bm{x}_{ll}^1, \bm{x}_{lh}^1, \bm{x}_{hl}^1, \bm{x}_{hh}^1 = DWT(\bm{x}),
\end{equation}
where $DWT(\cdot)$ refers to the DWT operation.
Despite that the input size is decreased by a factor of four ($H/2 \times W/2$) after a single DWT decomposition, DWT can be performed on the low-frequency component to further reduce the computational cost and expedite the inference:
\begin{equation}
     \bm{x}_{ll}^{J+1}, \bm{x}_{lh}^{J+1}, \bm{x}_{hl}^{J+1}, \bm{x}_{hh}^{J+1} = DWT(\bm{x}_{ll}^{J}),
\end{equation}
where $J$ is level of DWT decomposition and $\bm{x}_{ll}^{J+1},\bm{x}_{lh}^{J+1}, \bm{x}_{hl}^{J+1}, \bm{x}_{hh}^{J+1} \in \mathbb{R}^{{H}/{2^{(J+1)}} \times W/2^{(J+1)} \times 3}$.
Reversibly, given the frequency sub-bands, an image can be reconstructed via discrete inverse wavelet transform (IWT):
\begin{equation}
    \bm{x} = IWT(\bm{x}_{ll}^1, \bm{x}_{lh}^1, \bm{x}_{hl}^1, \bm{x}_{hh}^1).
\end{equation}

\noindent\textbf{Diffusion Models.}
Training of diffusion models (DMs) consists of a diffusion process and a denoising process.
Diffusion process transforms an image from the real data distribution $\bm{y}_0\sim q(\bm{y}_0)$ into a pure Gaussian noise $\bm{y}_T$ by successively applying the following Markov diffusion kernel:
\begin{equation}
    q(\bm{y}_t|\bm{y_{t-1}}) = \mathcal{N}(\bm{y}_t;\sqrt{1-\beta_t}\bm{y}_{t-1}, \beta_t\bm{I}),
\end{equation}
where $\{\beta_t\}_{t=1}^T$ is a pre-defined or learned noise variance schedule.
The marginal distribution at arbitrary timestep $t$ can be denoted as:
\begin{equation}
    q(\bm{y}_t|\bm{y}_0) = \mathcal{N}(\bm{y}_t;\sqrt{\alpha_t}\bm{y}_0, (1-\alpha_t)\bm{I}),
    \label{eq:diffuse}
\end{equation}
where $\alpha_t = \prod_{s=1}^t (1-\beta_s)$.
Given $\bm{y}_t$, the denoising process aims to recover $\bm{y}_0$ by recursively learning the transition from $\bm{y}_{t-1}$ to $\bm{y}_t$, which is defined as the following Gaussian distribution:
\begin{equation}
    p_{\theta}(\bm{y}_{t-1}|\bm{y}_t) = \mathcal{N}\left(\bm{y}_{t-1};\bm{\mu}_{\theta}(\bm{y}_t,t), \bm{\Sigma}_{\theta}(\bm{y}_t,t)\right),
    \label{eq:denoise}
\end{equation}
where parameters $\theta$ are optimized by a denoising network $\bm{\epsilon}_{\theta}$ that predicts $\bm{\mu}_{\theta}(\bm{y}_t,t)$ and $\mathbf{\Sigma}_{\theta}(\bm{y}_t,t)$.

\section{Methodology}
In the paper, \ours is proposed to handle blind face restoration (BFR) in the frequency domain.
The framework is depicted in \cref{fig:framework}.
First, degraded images are mapped to the frequency domain via Discrete Wavelet Transformation (DWT) (\cref{sec:dwt}).
The proposed Low-frequency Conditional Denoising (LCD) module (\cref{sec:LCD}) and High-Frequency Recovery (HFR) module (\cref{sec:HFR}) are adopted on the low- and high-frequency components respectively to remove degradations and restore facial details.
Recovered frequency components are used for the image reconstruction.

\subsection{Low-frequency Conditional Denoising}\label{sec:LCD}
As shown in \cref{fig:intro}, the low-frequency component is similar to the down-sampled version of the original image, which largely determines the restoration quality.
Due to its powerful generative ability from noisy inputs, a diffusion model (DM) is adopted in LCD to restore the low-frequency sub-band of high-quality (HQ) images.

\noindent \textbf{Conditional DM.}
We denote the low-frequency sub-band of a pair of LQ and HQ images as ($\bm{x}^j_{ll_0}$, $\bm{y}^j_{ll_0}$).
Since the low-frequency sub-band only contains a single map after DWT, the DWT levels $j$ and $ll$ are omitted in this section for simplicity.
According to the diffusion process defined in \cref{eq:diffuse}, images are successively destroyed by Gaussian noise as timestep $t$ increases, which means the distribution $q(\bm{x}_T|\bm{x}_0) \approx q(\bm{y}_T|\bm{y}_0) \approx \mathcal{N}(0, \bm{I})$ after a large timestep $T$.
Given this assumption, our LCD aims to learn the posterior distribution $p(\bm{y}_0 | \bm{x}_0)$:
\begin{equation}
    p(\bm{y}_0|\bm{x}_0) = \int q(\bm{y}_T|\bm{x}_0)\prod_{t=1}^T p_{\theta}(\bm{y}_{t-1}|\bm{y}_t)\mathrm{d}\bm{y}_{1:T},
    \label{eq:bfr_posterior}
\end{equation}
where $p_{\theta}(\bm{y}_{t-1}|\bm{y}_t)$ refers to the unconditional denoising process defined in \cref{eq:denoise}.
However, the unconditional scheme fails to preserve the original identity due to the lack of guidance during generation.
To solve the problem, the low-frequency sub-band of LQ images $\bm{x}_0$ are injected as the condition for the denoising process:
\begin{equation}
    p_{\theta}(\bm{y}_{t-1}|\bm{y}_t, \bm{x}_0) =
    \mathcal{N}\left(\bm{y}_{t-1};\bm{\mu}_{\theta}(\bm{y}_t,t,\bm{x}_0), \bm{\Sigma}_{\theta}(\bm{y}_t,t,\bm{x}_0)\right),
    \label{eq:cond_denoise}
\end{equation}
where $\bm{x}_0$ is injected by concatenating with the input $y_t$ along the channel dimension.

\noindent \textbf{Objectives.}
Following DDPM~\cite{ho2020denoising}, the denoising network $\bm{\epsilon}_{\theta}$ is trained to predict noise vectors with the objective:
\begin{equation}
    L_{LCD} = \mathbb{E}_{\bm{y}_0,t,\bm{\epsilon}_t\sim\N(\mathbf{0},\I)}\Big[\vert\vert\bm{\epsilon}_t - \bm{\epsilon}_{\theta}(\bm{y}_t,\bm{x}_0,t)\vert\vert^2 \Big].
    \label{eq:training_obj}
\end{equation}
During inference, with the learned parameterized Gaussian transitions $p_{\theta}(\bm{y}_{t-1} | \bm{y}_t, \bm{x}_0)$, the low-frequency sub-band of HQ images can be recovered from a random Gaussian noise $\bm{y}_T\sim\N(\mathbf{0},\I)$ by recursively applying:
\begin{equation}
    \bm{y}_{t-1}=\frac{1}{\sqrt{\alpha_t}}\left(\bm{y}_t - \frac{\beta_t}{\sqrt{1-\bar{\alpha}_t}} \bm{\epsilon}_{\theta}(\bm{y}_t,\bm{x}_0,t)\right) + \sigma_t\bm{z},
\end{equation}
where $\bm{z}\sim\mathcal{N}(\mathbf{0},\I)$, $\alpha_t=1-\beta_t$, and $\bar{\alpha}_t=\prod_{i=1}^t\alpha_i$.

%-------------------------------
\begin{figure*}
\centering
\begin{subfigure}{0.3\textwidth}
\includegraphics[height=2.8cm]{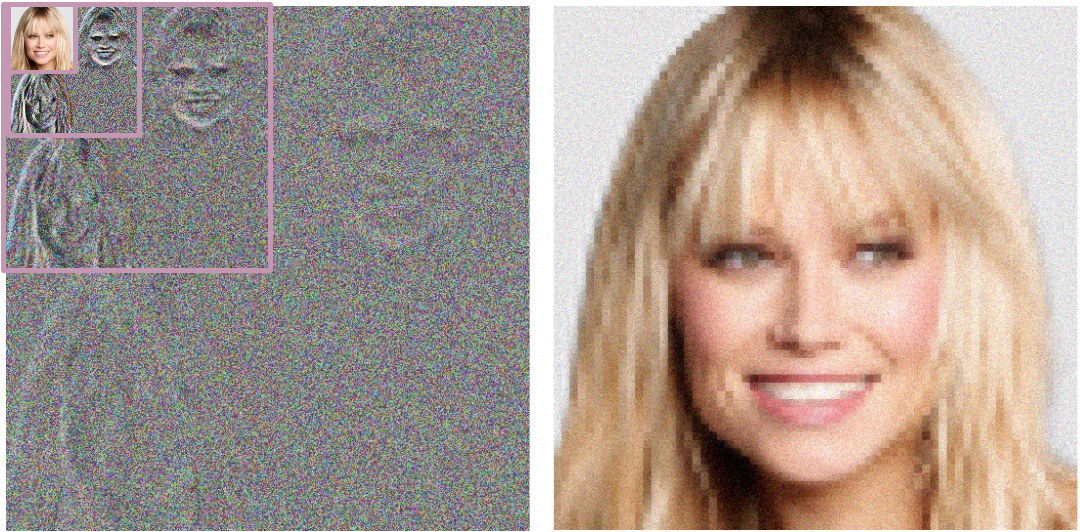} 
\caption{$J = 3 \quad (\times 64)$}
\end{subfigure}
\hspace{7mm}
\begin{subfigure}{0.3\textwidth}
\includegraphics[height=2.8cm]{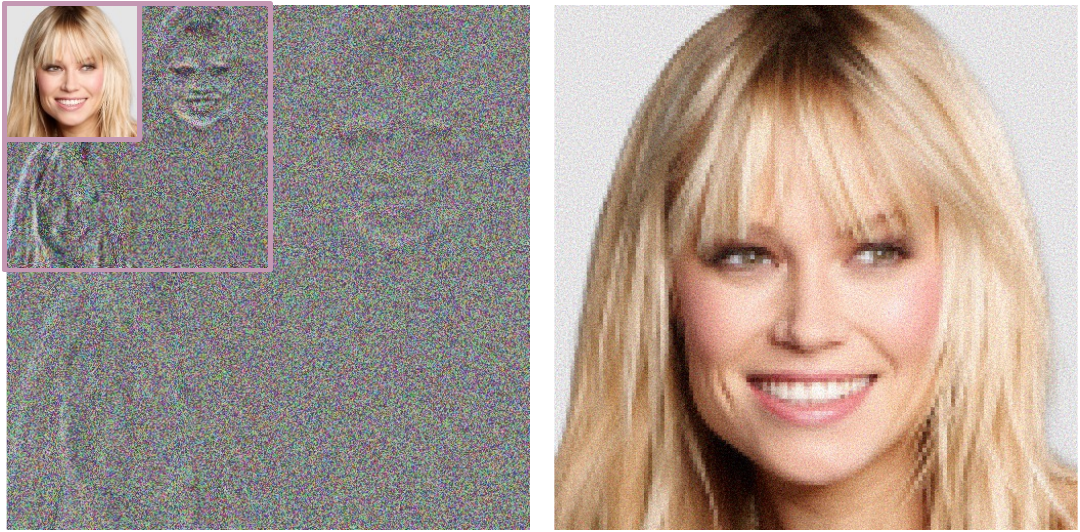} 
\caption{$J = 2 \quad (\times 128)$}
\end{subfigure}
\hspace{7mm}
\begin{subfigure}{0.3\textwidth}
\includegraphics[height=2.8cm]{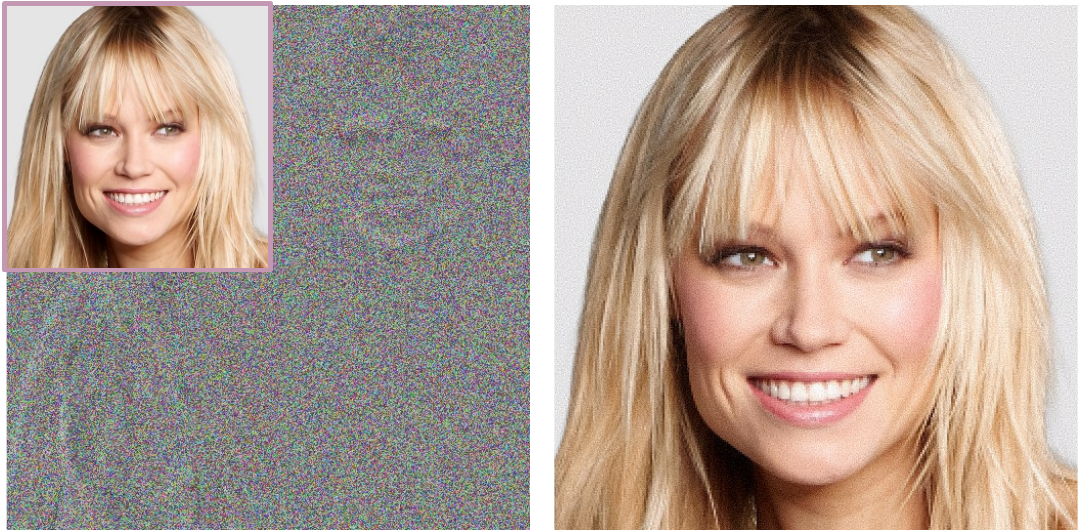} 
\caption{$J = 1 \quad (\times 256)$}
\end{subfigure}
\vspace{-5mm}
\caption{Visualization of DWT frequency components and images reconstructed by the low-frequency sub-band of an HQ image and high-frequency sub-bands of its LQ counterpart. DWT level $J$ and resolution of low- / high-frequency sub-bands ($\times N$) are reported.}
\vspace{-2mm}
\label{fig:vis_wave}
\end{figure*}
%-------------------------------
\noindent \textbf{Trade-off between efficiency and quality.}
Training a DM on large inputs ($\times$512 or more) requires massive computational resources and the evaluation takes thousands of steps to sample from the noise.
Although this problem can be alleviated by adopting low-frequency sub-bands at higher DWT levels, multiple times of decomposition will also lead to information reduction.
How to set DWT level wisely is the key to balance between efficiency and authenticity.

To illustrate how DWT level $J$ affects the authenticity, we visualize the image reconstructed by the low-frequency component of HQ images and the corresponding high-frequency ones of its LQ counterpart at different DWT levels in \cref{fig:vis_wave}.
Although noisy at the high-frequency part, the quality of images reconstructed at $J \leq 2$ is acceptable, with key facial features such as wrinkles still preserved.
However, the mosaic effect starts to emerge when $J$ keeps increasing.
Ablation studies about the effect of DWT level on training/inference are shown in \cref{sec:ab_dwt}.
In the paper, $J$ is set as 2 empirically.

\subsection{High-frequency Recovery}\label{sec:HFR}
Although the general face information is restored by LCD, rich facial details are generally embedded in the high-frequency component, as shown in \cref{fig:intro}, which improves the authenticity of restored images.
High-frequency sub-bands at multiple DWT levels also vary in size, which means more than one diffusion model is required for the recovery.
To avoid the extravagant computational cost, we adopt a U-shaped network that can recover high-frequency sub-bands at multiple DWT levels at the same time and takes only one step for inference.

%-------------------------------
\begin{figure}
    \centering
    \includegraphics[width=\linewidth]{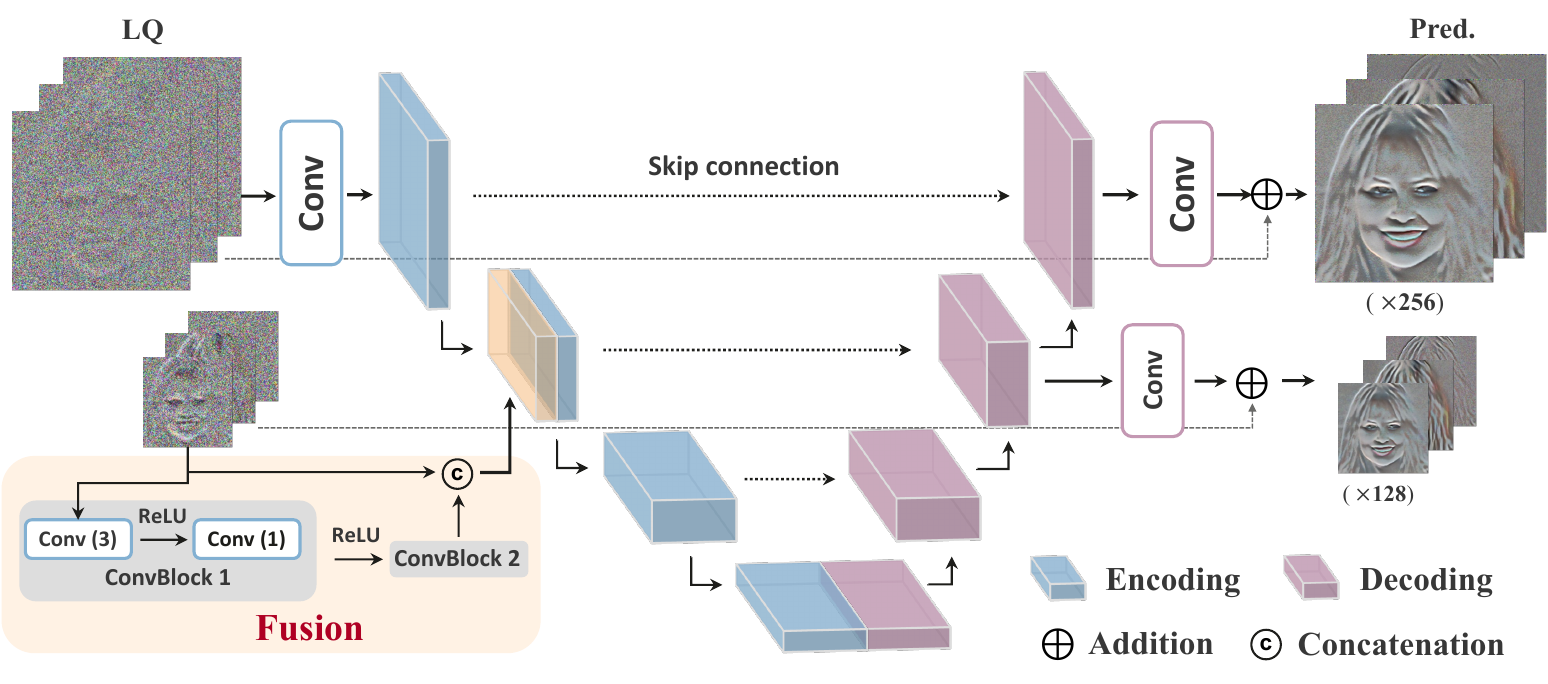}
    \caption{Illustration of high-frequency recovery (HFR) module.}
    \label{fig:hfr}
    \vspace{-3mm}
\end{figure}
%-------------------------------
Given a LQ and HQ image pair $(\bm{x}, \bm{y})$ and their high-frequency sub-bands: $\bm{x}^j_H = \{\bm{x}^j_{lh}, \bm{x}^j_{hl}, \bm{x}^j_{hh}\}$ and $\bm{y}^j_H = \{\bm{y}^j_{lh}, \bm{y}^j_{hl}, \bm{y}^j_{hh}\}$, where $j\in\{1,2\}$ refers to DWT level, the framework our high-frequency recovery (HFR) module is illustrated in \cref{fig:hfr}.
HFR takes high-frequency sub-bands of LQ images that are channel-wise concatenated as inputs and outputs the recovered ones $\bm{\hat{x}}^j_H$ as follows:
\begin{gather} \label{eq:hfr}
\bm{F}^1_H, \bm{F}^2_H = \mathcal{U}_{\phi_u}(\mathcal{H}_{\phi_{in}}(\bm{x}^1_H), \mathcal{F}_{\phi_f}(\bm{x}^2_H)), \\
\bm{\hat{x}}^j_H = \mathcal{H}_{\phi^j_{out}}(\bm{F}^j_H) + \bm{x}^j_H, \quad j\in \{1,2\},
\end{gather}
where $\mathcal{U}(\cdot)$, $\mathcal{H(\cdot)}$, $\mathcal{F}(\cdot)$ denotes the U-shaped network, the convolutional layer, and the fusion module with the parameter $\phi_u$, $\phi^j_{in/out}$, and $\phi_f$, respectively. 
$\bm{F}^1_H$, $\bm{F}^2_H$ represents feature maps of the last and penultimate decoding layers of the U-shaped network.
Considering efficiency, the fusion module only contains two stacked ConvBlocks that apply 3$\times$3 and 1$\times$1 convolutional layers with ReLU in between, followed by concatenation with the input.

\noindent \textbf{Objectives.}
We adopt two loss terms for the training of HFR: recovery loss $\mathcal{L}_r$ and content loss~\cite{wang2004ssim} $\mathcal{L}_c$. 
The recovery loss is applied not only to restore HQ high-frequency sub-bands $\bm{y}^j_H$ at each level but also to the image $\bm{y}^{j-1}$ reconstructed via inverse wavelet transform (IWT):
\begin{equation}
\mathcal{L}_r = \sum_j (||\bm{\hat{x}}^j_H - \bm{y}^j_H || + \alpha || IWT(\bm{y}^j_{ll}, \bm{\hat{x}}^j_H) - \bm{y}^{j-1}||), 
\end{equation}
where $IWT(\cdot)$ denotes IWT operation
and $\alpha=1$ is the parameter balancing losses between the frequency domain and the pixel domain.
In addition to minimizing the pixel-wise distance, the content loss is applied to maximize the luminance, contrast, and structural similarity of the restored and original HQ image:
\begin{equation}
\mathcal{L}_c = 1 - SSIM(IWT(\bm{y}^2_{ll}, \bm{\hat{x}}^2_H, \bm{\hat{x}}^1_H), \bm{y}).
\end{equation}
The overall objective of HFR is $\mathcal{L}_{HFR} = \mathcal{L}_r + \lambda \mathcal{L}_c$, where the weighting parameter $\lambda$ is set as 10.

%% file: sec/4_exper.tex
\section{Experiments}

\subsection{Settings and Datasets} \label{sec:dataset}

\noindent\textbf{Training Dataset.}
FFHQ~\cite{karras2019stylegan} is used as the training set, which contains 70,000 high-quality (HQ) face images. 
Following BFR benchmark works~\cite{wang2022restoreformer, gu2022vqfr}, we resize HQ images to the resolution of $512\times 512$, and then synthesize low-quality (LQ) counterparts as follows:
\begin{equation}
    \bm{x} = \left\{\left[\left(\bm{y} \otimes \bm{k}_{\sigma}\right){\downarrow_s} + \bm{n}_{\delta}\right]_{\text{JPEG}_q}\right\}{\uparrow_s},
    \label{eq:degradation}
\end{equation}
where a HQ image $\bm{y}$ is firstly blurred by a Gaussian kernel $\bm{k}_{\sigma}$, followed by a downsampling of scale $s$.
Afterward, Gaussian noise $\bm{n}_{\delta}$ and JPEG compression with quality factor $q$ are applied to the image, which is then upsampled back to the original size to obtain its LQ counterpart $\bm{x}$. 
The hyper-parameters $\sigma$, $s$, $\delta$, and $q$ are uniformly sampled from $[0.1,15]$, $[0.8,32]$, $[0,20]$, and $[30,95]$ respectively.

\noindent\textbf{Testing Dataset.}
We evaluate \ours on a synthetic dataset: CelebA-Test and three real-world datasets: LFW-Test~\cite{huang2008labeled}, WebPhoto-Test~\cite{wang2021gfpgan}, and WIDER-Test~\cite{zhou2022codeformer}.
CelebA-Test contains 3000 HQ images from CelebA-HQ~\cite{karras2018progressive}, and LQ counterparts are synthesized via Eq.~(\ref{eq:degradation}) with the same degradation setting.
In terms of three real-world datasets,
LFW-Test contains 1711 mildly degraded face images in the wild, which comprises the first image for each person in LFW~\cite{huang2008labeled}.
WebPhoto-Test includes 407 images crawled from the internet, some of which are old photos with severe degradation. 
WIDER-Test consists of 970 images with severe degradations from the WIDER dataset~\cite{yang2016wider}.

\noindent\textbf{Evaluation Metrics.}
For evaluation, we adopt two pixel-wise metrics (PSNR and SSIM), a reference perceptual metric (LPIPS~\cite{zhang2018lpips}), and a non-reference perceptual metric (FID~\cite{heusel2017fid}). 
To measure the consistency of identity, the angle between embeddings extracted by ArcFace (``Deg.'')~\cite{deng2019arcface} is used.
All metrics are used for the evaluation of synthetic data while only FID is used for real-world datasets due to the lack of referential HQ images.

\noindent\textbf{Implementation Details.}
The DWT decomposition level is set as 2, which reduces the input size of the diffusion model from 512$\times$512 to 128$\times$128.
We train LCD and HFR individually with Adam~\cite{kingma2015adam} optimizer.
LCD is trained for 200K iterations with the learning rate $1e\text{-}4$ and batch size 32.
For HFR, the training takes 70K iterations.
The learning rate gradually decays from $1e\text{-}3$ to $1e\text{-}5$ by a factor of 0.1.

\subsection{Ablation Studies}\label{sec:ab_study}
\subsubsection{Low-frequency Conditional Denoising} \label{sec:ab_dwt}
\noindent \textbf{DWT Levels.} 
We conduct an experiment to investigate how the level of wavelet decomposition (DWT) affects the efficiency and authenticity.
DWT is applied 0, 1, 2, 3 times on the training set, where the resolution of the diffusion model (DM)'s input becomes 512$\times$512, 256$\times$256, 128$\times$128, and 64$\times$64, respectively.
The number of parameters and the inference time are used to evaluate the efficiency while PSNR and SSIM are used to evaluate authenticity. 
The comparison results are reported in \cref{tab:res_comp}.
% ----------------------------------
\begin{table}
  \small
  \centering
    \caption{Comparison between state-of-the-art diffusion model-based BFR methods and ours at different DWT levels in terms of efficiency and authenticity. The best and the second best performances are \bred{highlighted} and \underline{underlined}.}
    \scalebox{0.71}{
        \begin{tabular}{c|c|c|c|c|c|c}
        \toprule
        \textbf{Method} & \textbf{Level} & \textbf{Resolution} & \textbf{\#Param. (M)} & \textbf{Time (s)} & \textbf{PSNR}$\uparrow$ & \textbf{SSIM}$\uparrow$ \\
        \midrule
        DifFace~\cite{yue2022difface} & - & $\times$ 512 & 175.38 & 25.04 & 19.06 & 0.46 \\
        DR2~\cite{wang2023dr2} & - & $\times$ 256 & 93.56 & 8.76 & 22.89 & 0.57\\
        \midrule
        \multirow{4}{*}{WaveFace} & 0 & $\times$ 512 & 109.06 & 19.37 & 26.42 & 0.65 \\
        & \bgcgrey{1} & \bgcgrey{$\times$ 256} & \bgcgrey{71.41} & \bgcgrey{8.12} & \bgcgrey{\bred{27.08}} & \bgcgrey{\bred{0.73}} \\
        & \bgcgrey{2} & \bgcgrey{$\times$ 128} & \bgcgrey{25.36} & \bgcgrey{1.97} & \bgcgrey{\underline{26.97}} & \bgcgrey{\underline{0.71}} \\
        & 3 & $\times$ 64 & 3.9 & 1.46 & 24.81 & 0.46 \\
        \bottomrule
  \end{tabular}}
\vspace{-0.4cm}
\label{tab:res_comp}
\end{table}
  % ----------------------------------
In comparison with the model trained on original images ($\times$512), our frequency-aware scheme applies DM on the low-frequency sub-band only, which avoids some unknown noise on high-frequency components.
Thus, the restoration quality improves after the wavelet decomposition is applied once.
The quality can be maintained until the DWT level increases to 2, where the inference process is accelerated by $10\times$.
However, the performance is significantly harmed when the DWT level further increases (3 or higher) as too many details in the low-frequency component have been reduced.
DWT level is set as 2 throughout the experiments to achieve the trade-off between the efficiency and the quality of restoration.

We also compare the efficiency with two DM-based BFR methods: DifFace~\cite{yue2022difface} and DR2~\cite{wang2023dr2}.
To make a fair comparison, the inference time refers to the time of the whole denoising process, which starts from the pure Gaussian noise.
The quality of generated images cannot compared with that reported in~\cref{tab:celeba_blind} as images are generated with pre-/post-processing and fewer sampling steps in their methods.
As can be seen, our method achieves better restoration quality at a lower computational cost.

\begin{figure}
    \centering
    \includegraphics[width=\linewidth]{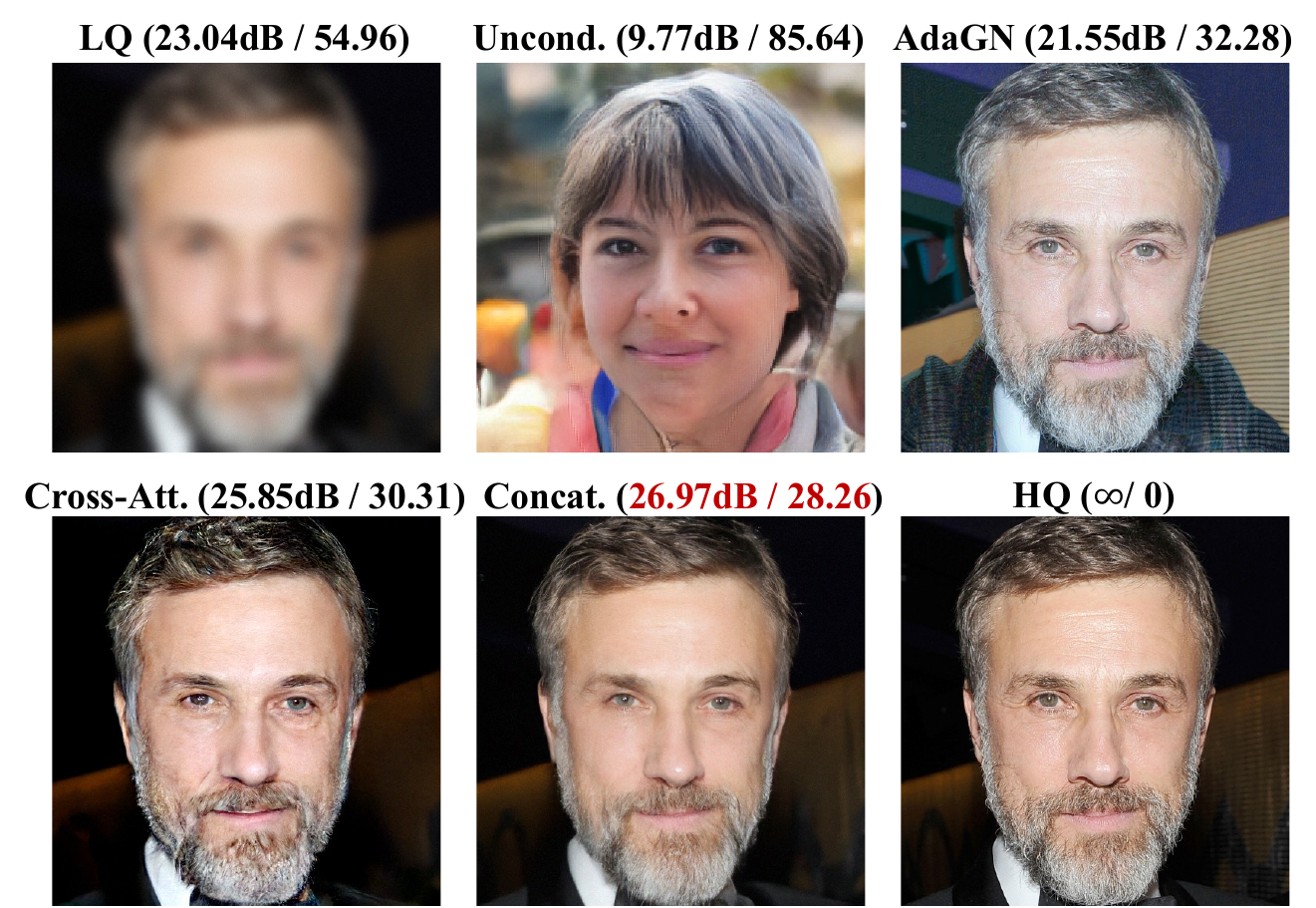}
    \caption{Qualitative comparison of different conditioning schemes on CelebA-Test. We adopt ``Concat.'' in \ours. PSNR($\uparrow$) / Deg.($\downarrow$) are reported.}
    \label{fig:ab_cond}
    \vspace{-4mm}
\end{figure}

\noindent \textbf{Conditions.}
To investigate how the condition facilitates the restoration, we first remove the condition from the input of our LCD module and train an unconditional diffusion model for 200K iterations for a fair comparison.
The restored face is denoted by ``Uncond.'' in \cref{fig:ab_cond}.
We notice that not only the identity is not preserved but also the image quality drops significantly.
This could be attributed to the slow convergence of the unconditional training manner~\cite{preechakul2022diffae}.

Additionally, we compare with two widely-used conditioning schemes: adaptive group normalization layers (AdaGN)~\cite{preechakul2022diffae} and cross-attention~\cite{rombach2022ldm}.
We replace the low-frequency sub-band of LQ images with the corresponding identity embedding extracted by Arcface~\cite{deng2019arcface}, which are subsequently injected via AdaGN.
Furthermore, a cross-attention layer is added to condition the denoised image at each step with the LQ low-frequency information.
Experimental details of the two schemes will be discussed in supplementary materials.
The restored images are shown in \cref{fig:ab_cond} as ``AdaGN'' and ``Cross-Att.'', respectively.
Although the original identity is preserved by the AdaGN scheme, the background is not well restored.
A possible reason is that the identity-related condition is performed at the latent level only, which fails to provide pixel-wise constraints.
On the other hand, despite the cross-attention scheme achieves a comparable restoration result with ours (``Concat.''), the extra layer requires more computational resources.
% In addition, the restored images are generally unnatural. For example, eyes are in different colors in \cref{fig:ab_cond}.
%-------------------------------
%-------------------------------
% ----------------------------------
\begin{table}[!htbp]
  \small
  \centering
    \caption{Ablation studies of the effectiveness of our high-frequency recovery (HFR) module. $\bm{x}_H$ denotes high-frequency sub-bands of LQ images and $1, 2$ stands for the level of wavelet decomposition. Best performance is indicated by \bred{Red}.}
    % \vspace{-2mm}
    \scalebox{0.84}{
        \begin{tabular}{c|c|c|c|c}
        \toprule
        & \textbf{PSNR}$\uparrow$ & \textbf{SSIM}$\uparrow$ & \textbf{LPIPS}$\downarrow$ & \textbf{FID}$\downarrow$ \\
        \midrule
        LCD + $\bm{x}^1_H$ + $\bm{x}^2_H$ & 24.353 & 0.430 & 0.537 & 17.264 \\
        LCD + HFR$_1$ + $\bm{x}^2_H$ & 26.081 & 0.609 & 0.481 & 14.020\\
        LCD + $\bm{x}^1_H$ + HFR$_2$& 24.819 & 0.460 & 0.536 & 15.293 \\
        LCD + HFR$_1$ + HFR$_2$ (Ours) & \bred{26.967} & \bred{0.711} & \bred{0.343} & \bred{13.062} \\
        \bottomrule
  \end{tabular}}
\vspace{-4mm}
\label{tab:ab_hfr}
\end{table}
% ----------------------------------
\subsubsection{High-frequency Recovery}
To investigate the effectiveness of our High-frequency Recovery (HFR) module, we replace the recovered high-frequency sub-bands at different DWT levels with that of low-quality (LQ) images for image reconstruction.
Quantitative comparisons on the restoration quality of different schemes are reported in \cref{tab:ab_hfr}.
The restoration quality drops significantly when HFR is removed with a decline of $2.62$ dB in PSNR.
Besides, the lack of recovered high-frequency components at different DWT levels deteriorates the restoration quality to varying extents.
In comparison with those at level 2 ($\times$128), high-frequency sub-bands at level 1 ($\times$256) contribute more to the restoration quality.

\begin{figure}
    \centering
    \includegraphics[width=1\linewidth]{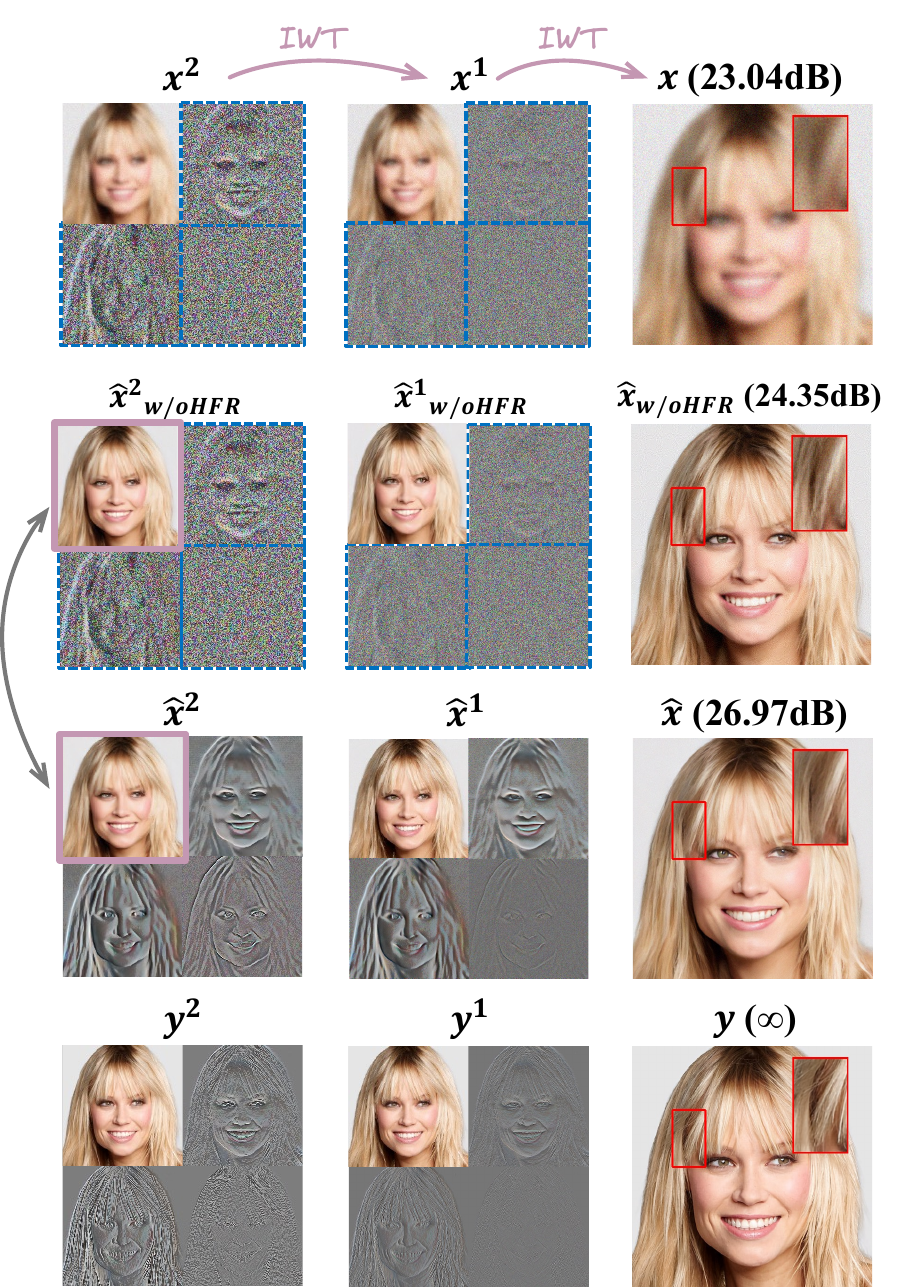}
    \caption{Qualitative comparison between images reconstructed at different DWT levels with and without (w/o) HFR module on CelebA-Test. $\bm{x}$, $\bm{\hat{x}}$ and $\bm{y}$ denotes LQ images, restored images, and HQ images. $\bm{x^j} (j\in\{1,2\})$ denotes frequency components at $j$-th DWT level. PSNR($\uparrow$) of the restored images are reported.}
    \label{fig:ab_hfr}
    \vspace{-5mm}
\end{figure}

The exemplar illustrated in \cref{fig:ab_hfr} helps to understand intuitively.
Before our HFR is applied (Row 2), the restored image contains cluttered noise and lacks details such as hairstyle due to messy high-frequency sub-bands of LQ images.
The quality improves when the image is constructed with the refined high-frequency component (Row 3), which is less noisy and presents fine-grained details.

%-------------------------------
%-------------------------------

% ----------------------------------
\begin{table}[h]
	\vspace{-3mm}
	\small
	\centering
	\caption{Quantitative comparison on \textbf{CelebA-Test} for blind face restoration. ``Deg.'' refers to the angle between identity embeddings of restored images and HQ counterparts. The best and the second best performances are \bred{highlighed} and \underline{underlined}.}
	\tabcolsep=0.1cm
	\scalebox{0.88}{
		% \hspace{-0.5cm}
		\begin{tabular}{c|cc|cc|c}
			\toprule
			\textbf{Methods} & \textbf{PSNR}$\uparrow$  & \textbf{SSIM}$\uparrow$ & \textbf{LPIPS}$\downarrow$  & \textbf{FID}$\downarrow$ & \textbf{Deg.}$\downarrow$    \\ 
            \midrule
			\bgcblue{Input} & \bgcblue{23.038} & \bgcblue{0.389}  & \bgcblue{0.586} & \bgcblue{67.505} & \bgcblue{54.958} \\
            \midrule
			GPEN~\cite{yang2021gpen} &  24.319  &  0.603 & 0.440 & 22.251 & 38.473\\
                GFP-GAN~\cite{wang2021gfpgan} &  24.771 & \underline{0.674} & 0.361 & 14.501 & 36.958\\
			VQFR~\cite{gu2022vqfr} & 23.735 &  0.614 & \underline{0.358} & 14.048 & 38.684\\
			RestoreFormer~\cite{wang2022restoreformer} & 24.191 & 0.636 &  0.364 & \bred{12.000} & \underline{34.752} \\
			CodeFormer~\cite{zhou2022codeformer} & \underline{25.071} & 0.672 &  0.359 & 14.385 & 37.677 \\
			DR2~\cite{wang2023dr2} &  23.583   &  0.613 &  0.402 &  14.671  & 51.030 \\
                DifFace~\cite{yue2022difface} &  24.155  &  0.667 &  0.390 & 13.379 & 49.958 \\
            \midrule
                \textbf{\ours} & \bred{26.967}\arrowpur{1.90} & \bred{0.711} &\bred{0.343}& \underline{13.062} & \bred{28.263}\arrowpur{6.49} \\ 
            \midrule
			% \bgcblue{GT} & \bgcblue{$\infty$}  & \bgcblue{1}  & \bgcblue{0}  & \bgcblue{0} & \bgcblue{0}\\ 
            \bottomrule
	\end{tabular}}
	\vspace{-0.2cm}
 \label{tab:celeba_blind}
\end{table}
% {\color{cyan}\textbf{\scriptsize{$\uparrow$}1.69}}
% ----------------------------------

%-------------------------------
\begin{figure*}[h]
\begin{subfigure}{0.123\textwidth}
\includegraphics[height=6cm]{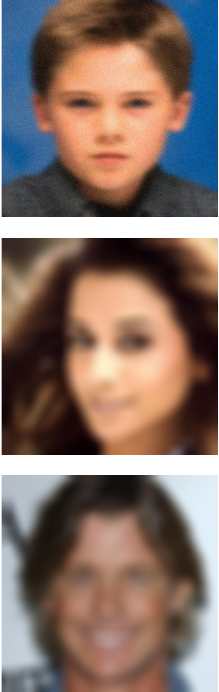} 
\subcaption*{LQ}
\end{subfigure}
\hspace{-4mm}
\begin{subfigure}{0.123\textwidth}
\includegraphics[height=6cm]{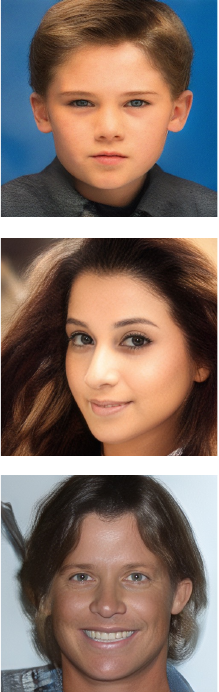} 
\subcaption*{GPEN~\cite{yang2021gpen}}
\end{subfigure}
\hspace{-4mm}
\begin{subfigure}{0.123\textwidth}
\includegraphics[height=6cm]{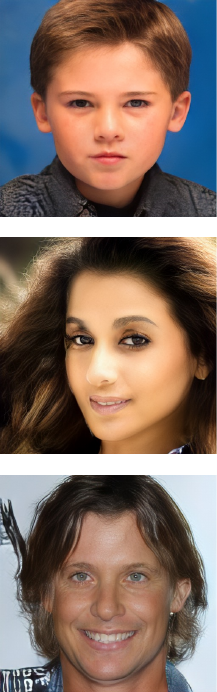} 
\subcaption*{GFPGAN~\cite{wang2021gfpgan}}
\end{subfigure}
\hspace{-4mm}
\begin{subfigure}{0.123\textwidth}
\includegraphics[height=6cm]{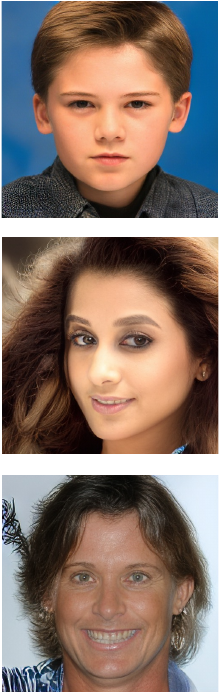} 
\subcaption*{VQFR\cite{gu2022vqfr}}
\end{subfigure}
\hspace{-4mm}
\begin{subfigure}{0.124\textwidth}
\includegraphics[height=6cm]{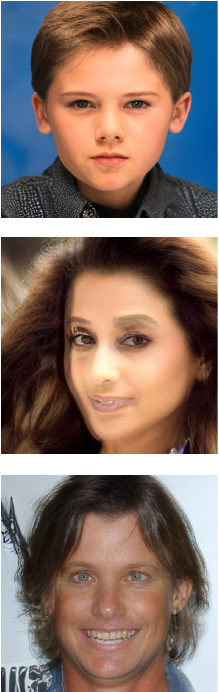} 
\subcaption*{\scriptsize{RestoreFormer}~\cite{wang2022restoreformer}}
\end{subfigure}
\hspace{-4mm}
\begin{subfigure}{0.123\textwidth}
\includegraphics[height=6cm]{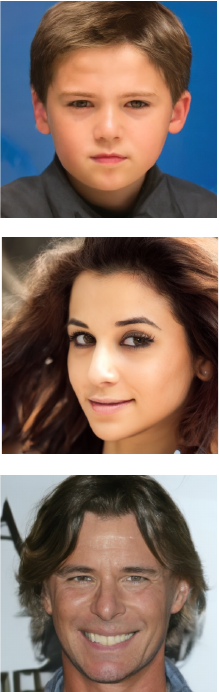} 
\subcaption*{DifFace~\cite{yue2022difface}}
\end{subfigure}
\hspace{-4mm}
\begin{subfigure}{0.123\textwidth}
\includegraphics[height=6cm]{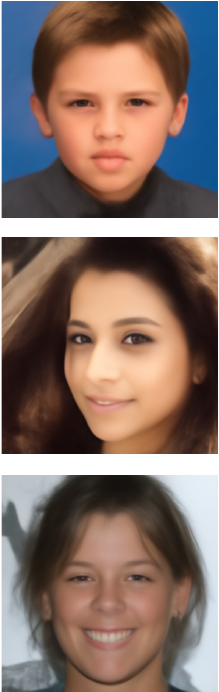} 
\subcaption*{DR2~\cite{wang2023dr2}}
\end{subfigure}
\hspace{-4mm}
\begin{subfigure}{0.123\textwidth}
\includegraphics[height=6cm]{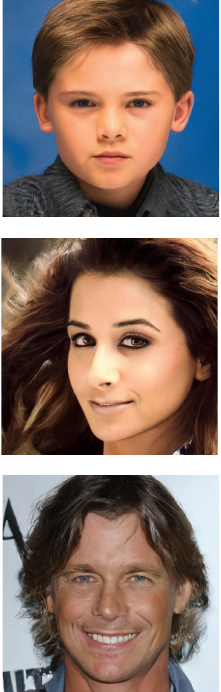} 
\subcaption*{\textbf{Ours}}
\end{subfigure}
\hspace{-4mm}
\begin{subfigure}{0.123\textwidth}
\includegraphics[height=6cm]{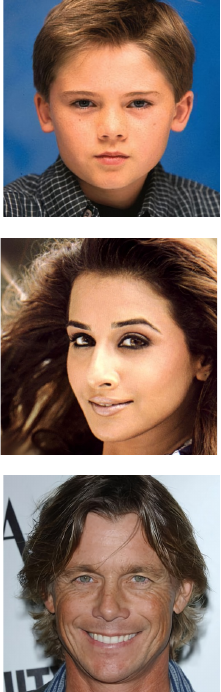 } 
\subcaption*{HQ}
\end{subfigure}
\hspace{-4mm}
\vspace{-3mm}
\caption{Qualitative comparison with state-of-the-art BFR methods on \textbf{CelebA-Test}. Our method achieves a better restoration quality with the original identity and facial details well preserved. (Zoom in for best view).}
\label{fig:syn}
\end{figure*}
%-------------------------------

\subsection{Comparisons with State-of-the-Art Methods}
State-of-the-art (SOTA) methods used for comparison include generative prior based methods (GPEN~\cite{yang2021gpen} and GFPGAN~\cite{wang2021gfpgan}), reference prior based methods (VQFR~\cite{gu2022vqfr} and RestoreFormer~\cite{wang2022restoreformer}) and recent diffusion model-based methods (DifFace~\cite{yue2022difface} and DR2~\cite{wang2023dr2}).
Evaluations are conducted on both synthetic and real-world datasets.

\noindent\textbf{Synthetic dataset.}
Quantitative comparison on CelebA-Test~\cite{karras2018progressive} are illustrated in~\cref{tab:celeba_blind}.
\ours achieves the best scores on reference-based metrics and ranks second on FID.
% \sout{Evaluation results on other non-reference metrics, such as NIQE~\cite{mittal2012niqe} and NRQM~\cite{ma2017nrqm}, are reported in supplementary materials.}
Apart from the better restoration quality, our method faithfully preserves the identity with the minimum angle between identity embeddings and their HQ counterparts, outperforming the second-best method by 6.5 degrees.

Additionally, we present the qualitative comparison with SOTA methods on images with increasing degradations in~\cref{fig:syn}.
GFPN and GFPGAN generally produce over-smoothed results.
VQFR and RestoreFormer introduce obvious artifacts especially when the inputs are corrupted by mild or severe degradation.
Besides, the above methods fail to preserve the identity since they are highly dependent on the StyleGAN prior, where the pre-defined latent space limits the diversity in restored results.
Diffusion prior-based methods (DifFace and DR2) tend to generate faithful images with minor artifacts. 
However, due to the lack of condition during generation, both methods still suffer from the failure in identity preservation.
% \sout{DR2 even mistakes the gender when encountering inputs with severe degradation.}
{On the contrary, our method, with the help of diffusion prior as well as the injected condition, is capable of generating high-quality facial images with fewer artifacts and, meanwhile, is faithful to the original identity.}

%-------------------------------
\begin{figure*}[h]
\begin{subfigure}{0.123\textwidth}
\includegraphics[height=6cm]{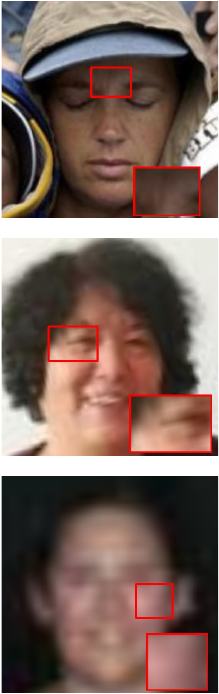} 
\subcaption*{LQ}
\end{subfigure}
\hspace{-4mm}
\begin{subfigure}{0.123\textwidth}
\includegraphics[height=6cm]{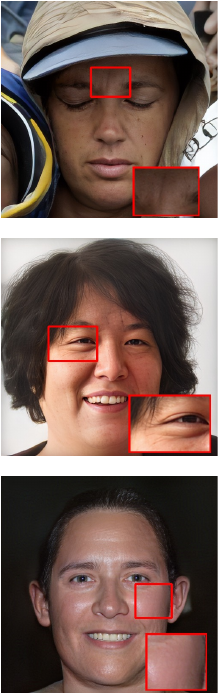} 
\subcaption*{GPEN~\cite{yang2021gpen}}
\end{subfigure}
\hspace{-4mm}
\begin{subfigure}{0.123\textwidth}
\includegraphics[height=6cm]{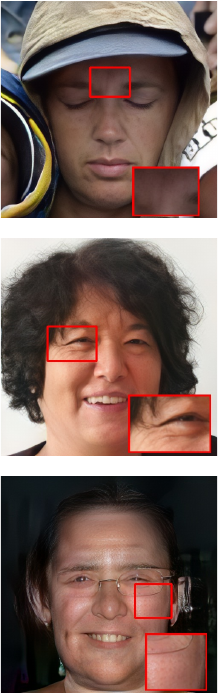} 
\subcaption*{GFPGAN~\cite{wang2021gfpgan}}
\end{subfigure}
\hspace{-4mm}
\begin{subfigure}{0.123\textwidth}
\includegraphics[height=6cm]{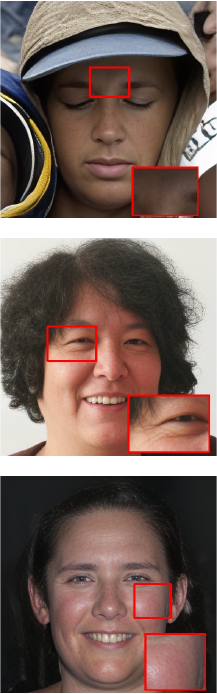} 
\subcaption*{VQFR\cite{gu2022vqfr}}
\end{subfigure}
\hspace{-4mm}
\begin{subfigure}{0.124\textwidth}
\includegraphics[height=6cm]{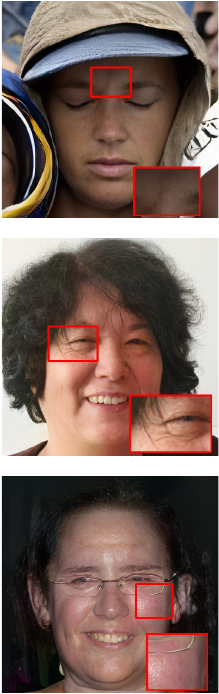} 
\subcaption*{\scriptsize{RestoreFormer}~\cite{wang2022restoreformer}}
\end{subfigure}
\hspace{-4mm}
\begin{subfigure}{0.123\textwidth}
\includegraphics[height=6cm]{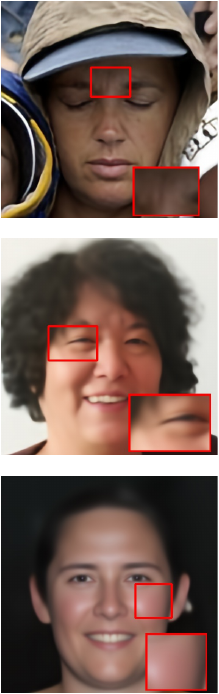} 
\subcaption*{SwinIR~\cite{liang2021swinir}}
\end{subfigure}
\hspace{-4mm}
\begin{subfigure}{0.123\textwidth}
\includegraphics[height=6cm]{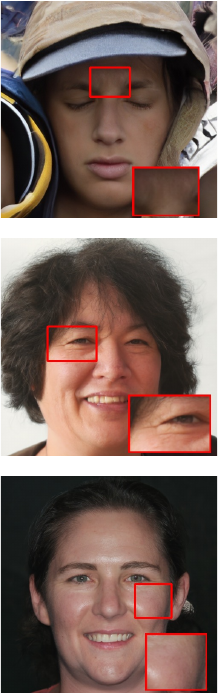} 
\subcaption*{DifFace~\cite{yue2022difface}}
\end{subfigure}
\hspace{-4mm}
\begin{subfigure}{0.123\textwidth}
\includegraphics[height=6cm]{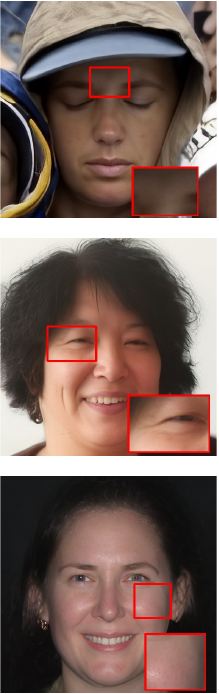} 
\subcaption*{DR2~\cite{wang2023dr2}}
\end{subfigure}
\hspace{-4mm}
\begin{subfigure}{0.123\textwidth}
\includegraphics[height=6cm]{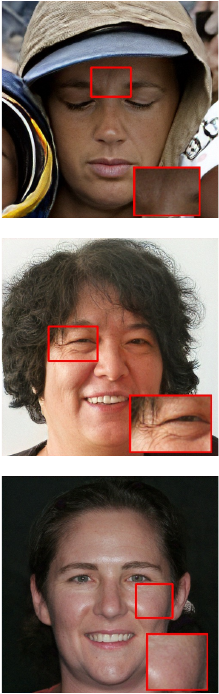} 
\subcaption*{\textbf{Ours}}
\end{subfigure}
\hspace{-4mm}
\vspace{-2mm}
\caption{Qualitative comparison with state-of-the-art BFR methods on \textbf{real-world} datasets, including LFW-Test (first row), WebPhoto-Test (second row), and WIDER-Test (third row). (Zoom in for best view).}
\vspace{-4mm}
\label{fig:real}
\end{figure*}
%-------------------------------
% ----------------------------------
\begin{table}
  \small
  \centering
    \caption{Quantitative comparisons on three \textbf{real-world datasets (-Test)} in terms of \textbf{FID}. The best and the second best performances are \bred{highlighted} and \underline{underlined}.}
    \scalebox{0.94}{
    \begin{tabular}{c|c|c|c}
    \toprule
    \textbf{Methods} & \textbf{LFW} & \textbf{WebPhoto} & \textbf{WIDER} \\    
    \midrule 
    \bgcblue{Input} & \bgcblue{124.974} & \bgcblue{170.112} & \bgcblue{199.961} \\
    \midrule
    GPEN~\cite{yang2021gpen} & 50.792 & 80.572 & 46.340 \\
    GFP-GAN~\cite{wang2021gfpgan} & 49.560 & 87.584  & 39.499 \\
    VQFR~\cite{gu2022vqfr} & 50.867&\bred{75.348}  & 44.107 \\
    RestoreFormer~\cite{wang2022restoreformer} &  47.750 &  \underline{77.330} & 49.817 \\
    CodeFormer~\cite{zhou2022codeformer} &  51.863 & 83.193 & 38.784 \\
    DR2~\cite{wang2023dr2} &  45.298 &  112.344 & 45.348  \\
    DifFace~\cite{yue2022difface} & \underline{45.227} & 87.811 & \underline{37.112} \\
    \midrule
    \textbf{\ours} & \bred{43.175} & 81.525 & \bred{36.913} \\ 
    \bottomrule
  \end{tabular}}
  \vspace{-3mm}
  \label{tab:real_fid}
\end{table}
% ----------------------------------

\noindent\textbf{Real-world datasets.} \label{sec:real_fid}
We compare with SOTA methods on three real-world datasets in terms of FID score, which measures the KL divergence between the distribution of restored images and that of HQ images in FFHQ~\cite{karras2019stylegan}.
{To reduce the large gap between the synthetic degradations and real-world ones, following previous diffusion prior-based methods~\cite{yue2022difface, liang2021swinir}, a pre-denoising network~\cite{liang2021swinir} is applied before our method to handle the unknown degradations.}
Quantitative results with SOTA methods are reported in~\cref{tab:real_fid}.
\ours outperforms state-of-the-art (SOTA) methods on LFW-Test~\cite{huang2008labeled} and WIDER-Test~\cite{zhou2022codeformer} and beats previous diffusion model based BFR methods on WebPhoto-Test~\cite{wang2021gfpgan}.
As mentioned in DifFace~\cite{yue2022difface}, the FID score may not be the most suitable evaluation metric for WebPhoto-Test, as a total of 407 images is significantly distant from representing the data distribution.

Three typical examples are shown in~\cref{fig:real} with the increasing degradation level.
It is observed that previous BFR methods tend to generate either hazy or unnatural faces.
Some methods even fail to generate adequate restored results when handling severe degradations (Row 3).
In contrast, our approach provides much more natural and realistic results with rich details such as wrinkles.

%% file: sec/5_con.tex
\section{Conclusion}
\vspace{-2mm}
We propose \ours to solve the task in the frequency domain to achieve the trade-off between efficiency and authenticity.
Efficiency is achieved by applying a diffusion model only on the low-frequency sub-band whose size is 1/16 of the original image.
Meanwhile, high-frequency components decomposed at multiple DWT levels are recovered simultaneously by a one-forward framework, which ensures the preservation of facial details.

\noindent \textbf{Limitations.}
There is a considerable gap between real-world degradations and the simulated ones (\cref{eq:degradation}), leading to inferior results in some cases.
Our future work will focus on how to simulate degradations that fit real-world scenarios.

%% file: sec/X_suppl.tex
\clearpage
\setcounter{page}{1}
\maketitlesupplementary

\renewcommand{\thesection}{\Alph{section}}

\counterwithin{table}{section}
\counterwithin{figure}{section}
\setcounter{table}{0}
\setcounter{section}{0}
\setcounter{figure}{0}

\section{Experimental Details}
\subsection{Conditioning scheme}
In~\cref{sec:ab_study}, we discuss the effectiveness of several widely-used conditioning schemes.
In this section, we will provide more details about ``AdaGN'' and ``Cross-Att.''.

\noindent \textbf{AdaGN.} 
Adaptive group normalization (AdaGN) conditions the denoising network at the latent level, where the timestep and latent vector are incorporated into each residual block after group normalization:
\begin{equation}
    \text{AdaGN}(\bm{h}, t, \bm{z}_{ll}) = \bm{z}_{s}(t_s \text{GroupNorm}(h) + t_b),
\end{equation}
where $\bm{z}_s \in \mathbb{R}^d = \text{Affine}(\bm{z}_{ll})$ refers to identity features of LQ images $\bm{x}_0$ extracted by ArcFace~\cite{deng2019arcface} after the affine transformation.
$(t_s, t_b) \in \mathbb{R}^{2 \times d} = MLP(\psi(t))$ is the output of a Multi-Layer Perceptron (MLP) with a sinusoidal encoding function $\psi$.
$d$ denotes the dimension of embeddings.

\noindent \textbf{Cross-Att.}
Cross-attention (CA) layer can improve the model performance via the inner relationship between inputs from multiple modalities~\cite{dhariwal2021diffusion,rombach2022ldm}.
In the paper, cross-attention is adopted to complement the denoised HQ sample at each timestep $\bm{y}_t$ with its LQ counterpart $\bm{x}_0$:
\begin{gather}
\text{CA}(Q, K, V) = \text{softmax}\left(\frac{QK^T}{\sqrt{d}}\right) \cdot V, \\
Q = W_Q \cdot \bm{x}_0, \quad K = W_K \cdot \bm{y}_t, \quad V = W_V \cdot \bm{y}_t,
\end{gather}
where $W_Q, W_K, W_V$ are learnable projection matrices~\cite{vaswani2017attention}.
$\text{CA}(\cdot)$ refers to the cross-attention operation, which is the pixel-wise dot product between HQ feature maps $V$ and corresponding attention scores.

\section{Additional Experimental Results}

% ----------------------------------
\begin{table}
  \small
  \centering
  \scalebox{0.9}{
  \begin{tabular}{c|c|c|c|c}
    \toprule
    \textbf{Methods} & \textbf{CelebA}& \textbf{LFW} & \textbf{WebPhoto} & \textbf{WIDER} \\    
    \midrule 
    \bgcblue{Input} & \bgcblue{14.114} & \bgcblue{8.575} & \bgcblue{12.664} & \bgcblue{13.498} \\
    \midrule
    GPEN~\cite{yang2021gpen} & 7.760 & 3.853 & 4.498 & 4.105 \\
    GFP-GAN~\cite{wang2021gfpgan} & 4.171 & 3.954 & 4.248 & 3.880 \\
    VQFR~\cite{gu2022vqfr} &  3.775 &  3.574  &  3.606  & 3.054 \\
    RestoreFormer~\cite{wang2022restoreformer} &  4.436 & 4.145 &  4.459  & 3.894 \\
    CodeFormer~\cite{zhou2022codeformer} &  4.680   & 4.520 & 4.708  &  4.165  \\
    DR2~\cite{wang2023dr2}    &  4.998   & 4.736  &  6.159 &  5.171 \\
    DifFace~\cite{yue2022difface}  & 4.500  & 4.220  &  4.666 & 4.688  \\
    \midrule
    \textbf{DM} & 4.898 & 4.784 & 4.860 & 4.988\\
    \textbf{\ours} & 4.421  &  4.133  & 4.383 & 4.963 \\ 
    \midrule 
    \bgcblue{GT} & \bgcblue{4.373} & \bgcblue{-} & \bgcblue{-} & \bgcblue{-} \\
    \bottomrule
  \end{tabular}}
  % \vspace{-0.1cm}
  \caption{Quantitative comparisons on synthetic and real-world datasets (-Test) in terms of \textbf{NIQE}$\downarrow$.}
  \label{tab:niqe}
\end{table}
% ----------------------------------
\subsection{Discussion on Non-Reference Metric}
In~\cref{sec:real_fid}, we report the performance of state-of-the-art (SOTA) methods on synthetic and real-world datasets in terms of FID score.
In this section, we further consider two commonly used non-reference metrics: NIQE~\cite{mittal2012niqe} and NRQM~\cite{ma2017nrqm}.
The results are reported in \cref{tab:niqe} and \cref{tab:nrqm}, respectively.
As can be seen, despite the superiority of our method in terms of identity preserving and facial detail recovery, it shows inferior performance on these non-reference metrics.
We also notice that images restored by some methods even unreasonably beat ground truth (GT) on CelebA-Test.
% ----------------------------------
\begin{table}
  \small
  \centering
  \scalebox{0.9}{
  \begin{tabular}{c|c|c|c|c}
    \toprule
    \textbf{Methods} & \textbf{CelebA}& \textbf{LFW} & \textbf{WebPhoto} & \textbf{WIDER} \\
    \midrule 
    \bgcblue{Input} & \bgcblue{6.042} & \bgcblue{2.810} & \bgcblue{2.044} & \bgcblue{1.358} \\
    \midrule
    GPEN~\cite{yang2021gpen} & 8.514 & 8.482 & 7.584 & 8.112 \\
    GFP-GAN~\cite{wang2021gfpgan} & 7.985 & 7.782 & 7.750 & 7.990 \\
    VQFR~\cite{gu2022vqfr} & 8.657 & 8.564  &  8.457  & 8.792 \\
    RestoreFormer~\cite{wang2022restoreformer} &  8.495   & 8.572 & 8.133 & 8.537 \\
    CodeFormer~\cite{zhou2022codeformer} &  8.339   &  8.217  & 7.457  & 8.370  \\
    DR2~\cite{wang2023dr2} &  6.906 & 6.049  &  4.423   & 5.219  \\
    DifFace~\cite{yue2022difface} & 7.724 & 6.322  & 4.929 & 4.728 \\
    \midrule
    \textbf{DM} & 7.121 & 7.125 & 7.091 & 7.167 \\
    \textbf{\ours} & 7.732  & 7.753  & 6.749 & 6.541\\ 
    \midrule
    \bgcblue{GT} & \bgcblue{7.909} & \bgcblue{-} & \bgcblue{-} & \bgcblue{-} \\
    \bottomrule
  \end{tabular}}
  \caption{Quantitative comparisons on synthetic and real-world datasets (-Test) in terms of \textbf{NRQM}$\uparrow$.} 
  \label{tab:nrqm}
\end{table}
% ----------------------------------

% -------------------------------
\begin{figure*}[t]
\centering
\includegraphics[height=4cm]{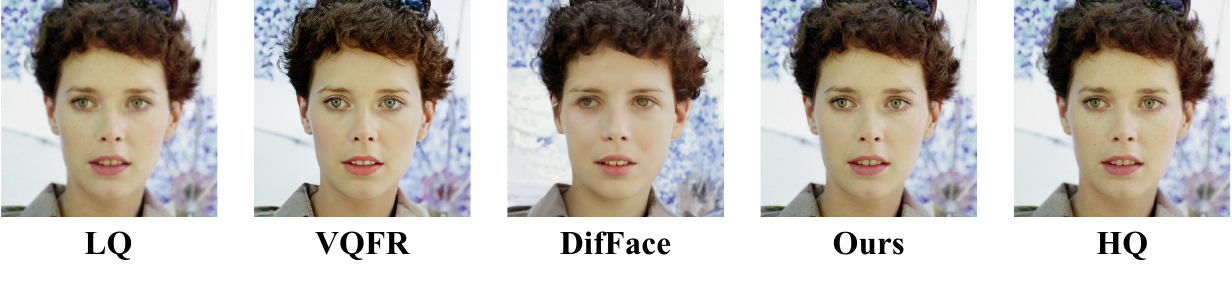}
\caption{Qualitative comparison between generative prior-based and diffusion model-based methods.}
\label{fig:fake}
\end{figure*}
% -------------------------------

% -------------------------------
\begin{figure*}
\centering
\includegraphics[width=\textwidth]{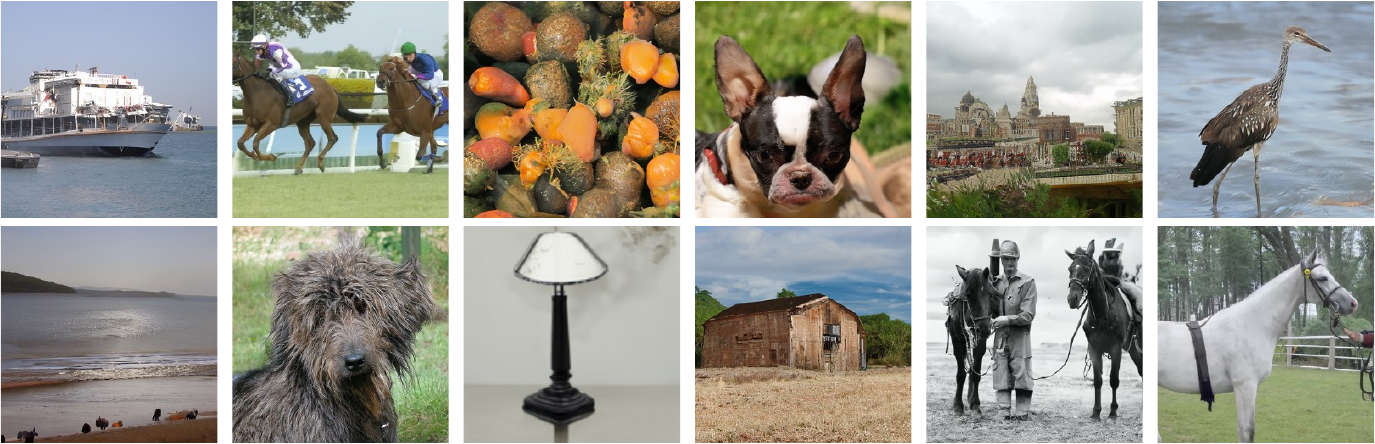}
\vspace{-2mm}
\caption{Samples generated by benchmark diffusion model.}
\label{fig:gd_examples}
\end{figure*}
% -------------------------------

To explore the reason behind this, the qualitative comparison is conducted between a generative prior-based method (VQFR~\cite{gu2022vqfr}) and diffusion model-based methods (DifFace~\cite{yue2022difface} and ours).
As shown in \cref{fig:fake}, although the image generated by VQFR provides better sharpness, it contains many artifacts.
For example, the hair and eyelashes present an unnatural woolen texture, which deteriorates the image's authenticity.
On the contrary, diffusion model-based methods can yield more photorealistic faces with human-like textures.

To further investigate the performance of diffusion models on these two metrics, we randomly sample the same number of images as the corresponding dataset with a benchmark pre-trained diffusion model\footnote{\url{https://github.com/openai/improved-diffusion}} and evaluate the quality by NIQE and NRQM.
Some generated images are depicted in \cref{fig:gd_examples} and quantitative results are denoted as ``DM'' in \cref{tab:niqe} and \cref{tab:nrqm}.
As shown in both tables, even images generated by the benchmark diffusion model underperform those by GAN-based methods on these non-reference metrics, which challenges the common finding that diffusion models beat GANs in image synthesis~\cite{dhariwal2021diffusion,rombach2022ldm,phung2023wavediff}.

Both qualitative and quantitative results illustrate that these two non-reference metrics cannot well represent the performance of BFR methods.
More investigations are called to study the appropriate evaluation metrics for BFR.

%-------------------------------
\begin{figure*}
\centering
\includegraphics[width=\textwidth]{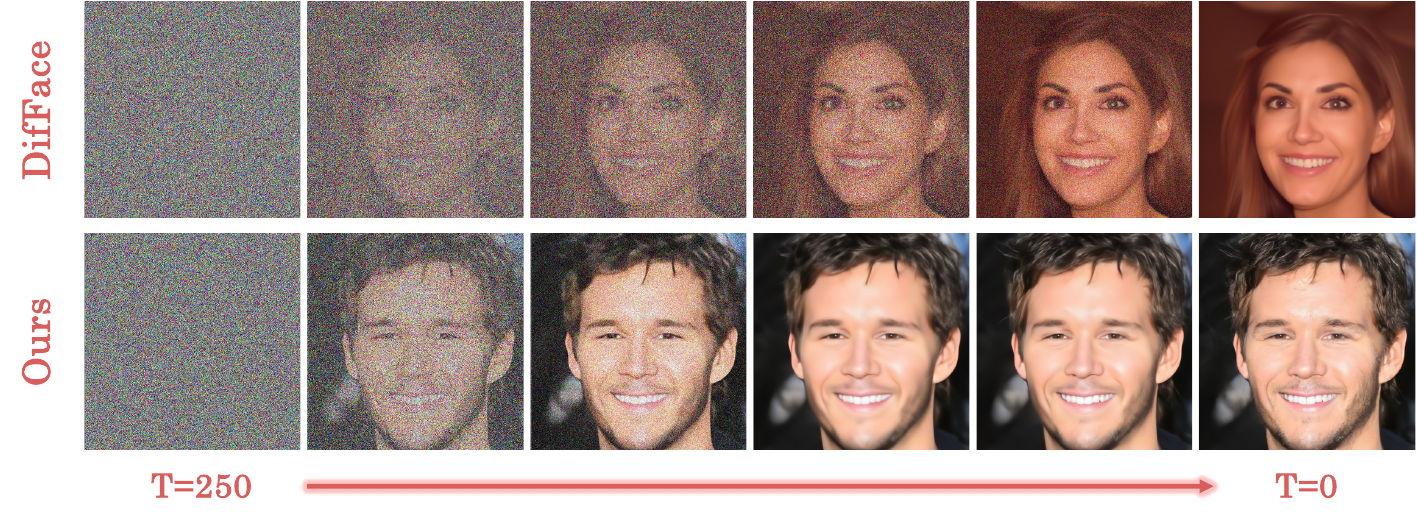}
\caption{Visualization of denoising process of the unconditional diffusion model adopted in DifFace~\cite{yue2022difface} and our conditional one.}
\vspace{-2mm}
\label{fig:denoising_process}
\end{figure*}
%-------------------------------

\subsection{Degradation types}
Apart from the classical degradation model (\cref{eq:degradation}), we adopt the second-order degradation process proposed by RealESRGAN~\cite{wang2021realesrgan}, where classical degradations are applied repeatedly to mimic real-world degradation.
Following the settings in~\cref{sec:dataset}, we train both LCD and HFR modules on FFHQ~\cite{karras2019stylegan} with RealESRGAN degradations.
To evaluate the model, a corresponding evaluation set is synthesized based on 3000 CelebA-HQ images, namely CelebA-Test-RESR.

\begin{figure*}
\centering
\includegraphics[width=\textwidth]{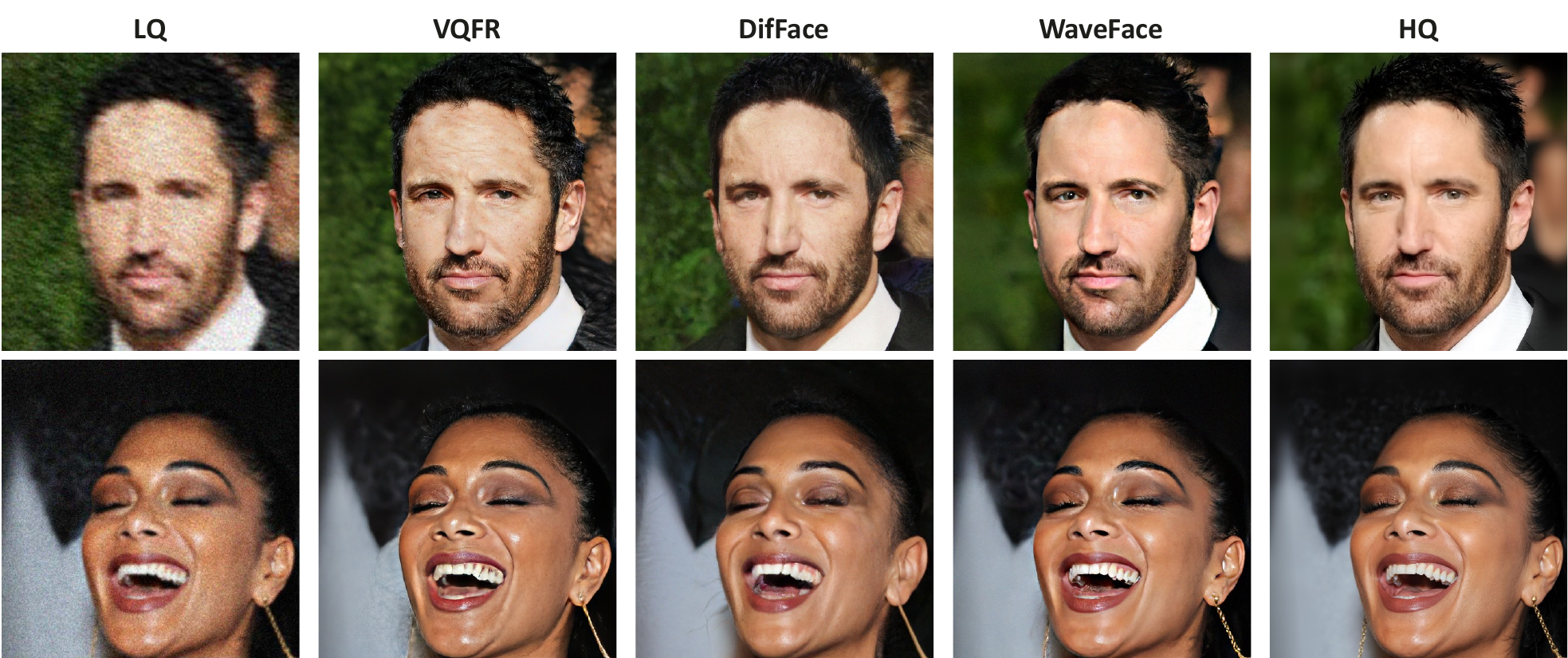}
\caption{Qualitative comparison on CelebA-Test-RESR.}
\vspace{-2mm}
\label{fig:esr}
\end{figure*}

The performances of SOTA methods and our method on CelebA-Test-RESR and real-world datasets are reported in \cref{tab:realesrgan} and \cref{tab:esr_real}, respectively.
As can be seen, the model trained on data with RealESRGAN degradations outperforms that on classical degradations, which demonstrates that RealESRGAN degradations can better imitate real-world degradations.
The qualitative comparison between SOTA methods and ours is illustrated in \cref{fig:esr}.
Our method (WaveFace) is able to deliver authentic results with both identity information and fine-grained details well preserved.
For example, our method restores more details of the earrings in 2nd column.

% ----------------------------------
\begin{table}[h]
	\vspace{-3mm}
	\small
	\centering
	\caption{Quantitative comparison on \textbf{CelebA-Test-RESR} for blind face restoration. ``Deg.'' refers to the angle between identity embeddings of restored images and HQ counterparts. Best performances are \bred{highlighed}.}
	\tabcolsep=0.1cm
	\scalebox{0.88}{
		% \hspace{-0.5cm}
		\begin{tabular}{c|cc|cc|c}
			\toprule
			\textbf{Methods} & \textbf{PSNR}$\uparrow$  & \textbf{SSIM}$\uparrow$ & \textbf{LPIPS}$\downarrow$  & \textbf{FID}$\downarrow$ & \textbf{Deg.}$\downarrow$    \\ 
            \midrule
			\bgcblue{Input} & \bgcblue{18.886} & \bgcblue{0.449}  & \bgcblue{0.574} & \bgcblue{48.968} & \bgcblue{39.910} \\
            \midrule
			VQFR~\cite{gu2022vqfr} & 18.167 & 0.516 & 0.459 & 11.911 & 35.104\\
            DifFace~\cite{yue2022difface} & 18.321 & 0.540 & 0.489 & 12.353 & 43.773 \\
            \midrule
            \textbf{\ours} & \bred{19.126} & \bred{0.576} & \bred{0.436} & \bred{11.336} & \bred{32.863}\\ 
            \midrule
            \bottomrule
	\end{tabular}}
	\vspace{-0.2cm}
 \label{tab:realesrgan}
\end{table}
% ----------------------------------

% ----------------------------------
\begin{table}
  \small
  \centering
    \caption{Quantitative comparisons on three \textbf{real-world datasets (-Test)} in terms of \textbf{FID$_\downarrow$}. Best performances are \bred{highlighed}.}
    \scalebox{0.94}{
    \begin{tabular}{c|c|c|c}
    \toprule
    \textbf{Methods} & \textbf{LFW} & \textbf{WebPhoto} & \textbf{WIDER} \\    
    \midrule 
    \bgcblue{Input} & \bgcblue{124.974} & \bgcblue{170.112} & \bgcblue{199.961} \\
    \midrule
    \textbf{\ours (Classical)} & 43.175 & 81.525 & 36.913 \\ 
    \textbf{\ours (RealESRGAN)} & \bred{46.711} & \bred{78.438} & \bred{35.750} \\ 
    \bottomrule
  \end{tabular}}
  \vspace{-3mm}
  \label{tab:esr_real}
\end{table}
% ----------------------------------

%-------------------------------
%-------------------------------

\subsection{Denoising Process Visualization}
We illustrate the denoising process of the diffusion model used in our low-frequency conditional denoising (LCD) module and the unconditional one adopted in DifFace~\cite{yue2022difface} in \cref{fig:denoising_process}.
Both models are trained on FFHQ~\cite{karras2019stylegan} and tested on CelebA-Test~\cite{karras2018progressive}.
We take DDIM~\cite{song2020denoising} as the sampling scheme, which takes 250 steps to sample an image from a pure Gaussian noise.
It can be easily observed that conditional DM (Ours) achieves a faster sampling convergence due to the prior knowledge provided by LQ counterparts.

\subsection{More Qualitative Comparisons}
\begin{figure*}
\begin{subfigure}{0.123\textwidth}
\includegraphics[height=12cm]{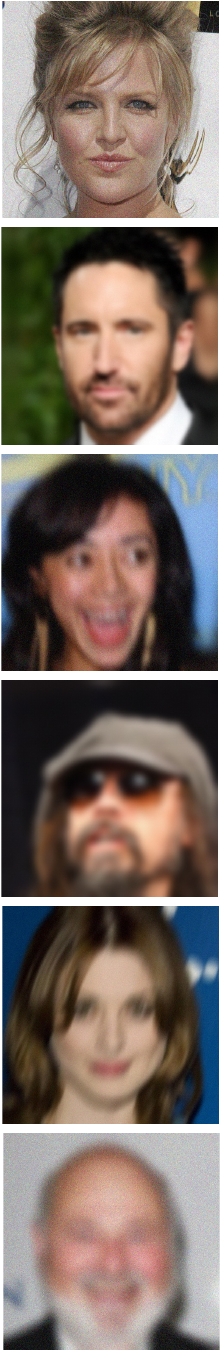} 
\subcaption*{LQ}
\end{subfigure}
\hspace{-3mm}
\begin{subfigure}{0.123\textwidth}
\includegraphics[height=12cm]{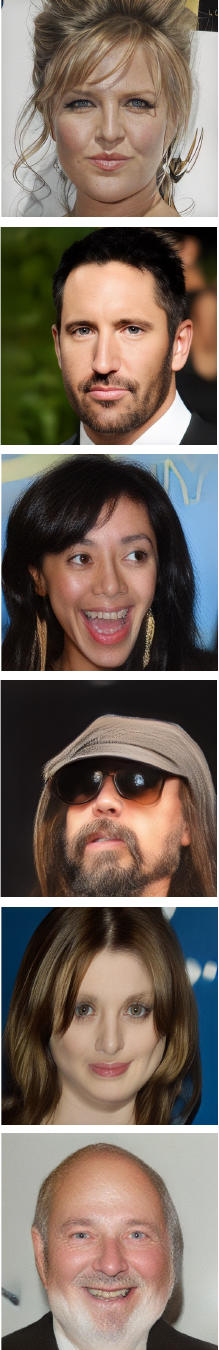} 
\subcaption*{GPEN~\cite{yang2021gpen}}
\end{subfigure}
\hspace{-4mm}
\begin{subfigure}{0.123\textwidth}
\includegraphics[height=12cm]{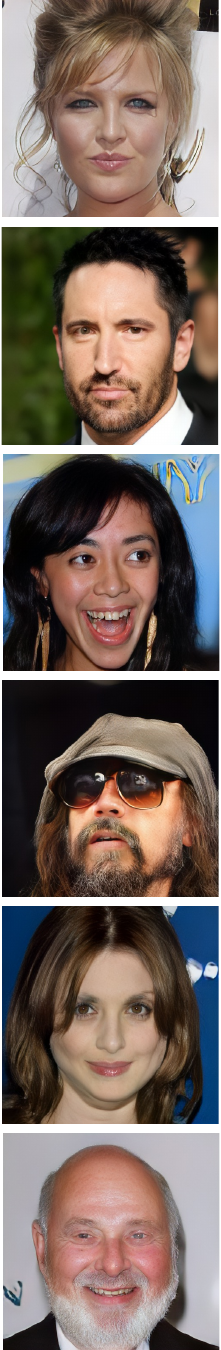} 
\subcaption*{GFPGAN~\cite{wang2021gfpgan}}
\end{subfigure}
\hspace{-4mm}
\begin{subfigure}{0.123\textwidth}
\includegraphics[height=12cm]{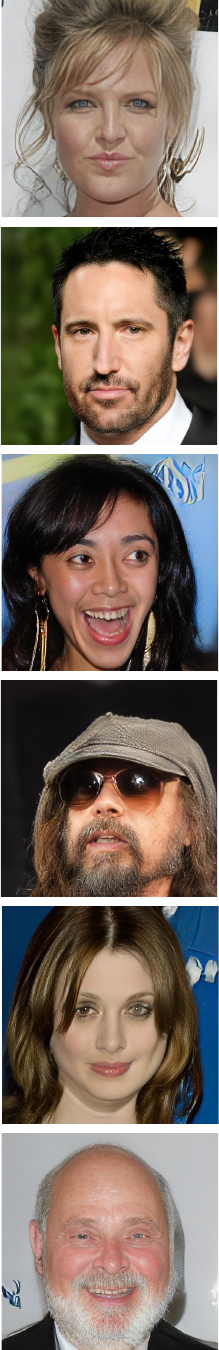} 
\subcaption*{VQFR\cite{gu2022vqfr}}
\end{subfigure}
\hspace{-3.8mm}
\begin{subfigure}{0.124\textwidth}
\includegraphics[height=12cm]{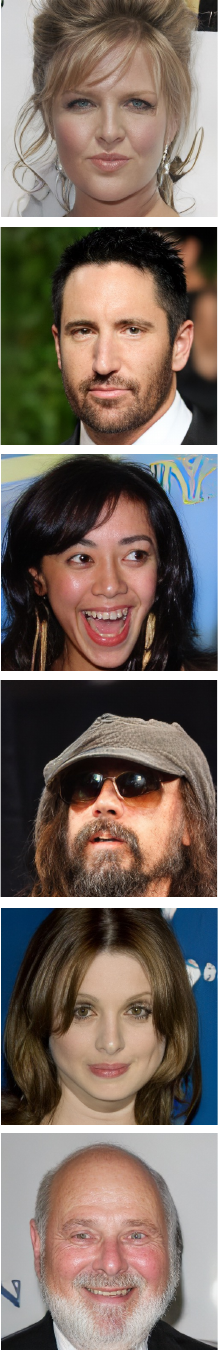} 
\subcaption*{\scriptsize{RestoreFormer}~\cite{wang2022restoreformer}}
\end{subfigure}
\hspace{-3.8mm}
\begin{subfigure}{0.123\textwidth}
\includegraphics[height=12cm]{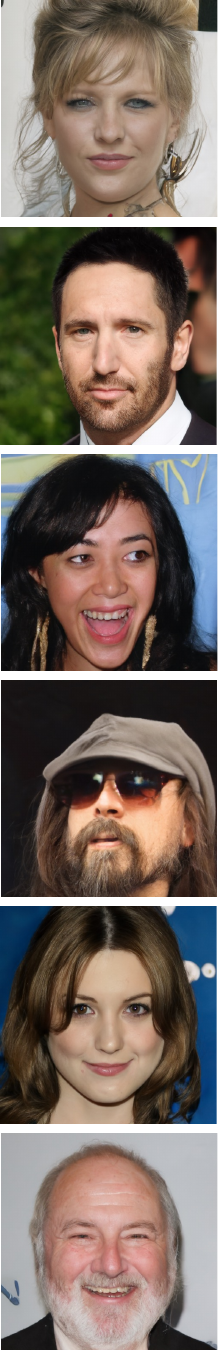} 
\subcaption*{DifFace~\cite{yue2022difface}}
\end{subfigure}
\hspace{-4mm}
\begin{subfigure}{0.123\textwidth}
\includegraphics[height=12cm]{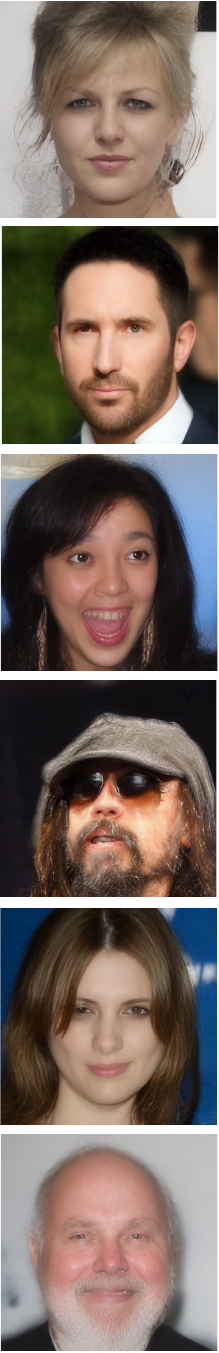} 
\subcaption*{DR2~\cite{wang2023dr2}}
\end{subfigure}
\hspace{-4mm}
\begin{subfigure}{0.123\textwidth}
\includegraphics[height=12cm]{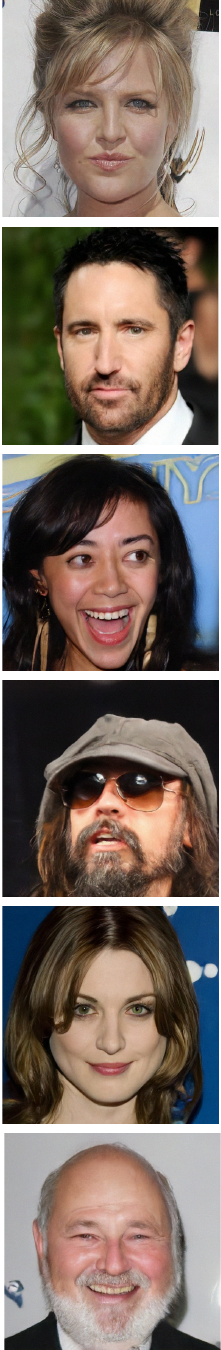} 
\subcaption*{\textbf{Ours}}
\end{subfigure}
\hspace{-3mm}
\begin{subfigure}{0.123\textwidth}
\includegraphics[height=12cm]{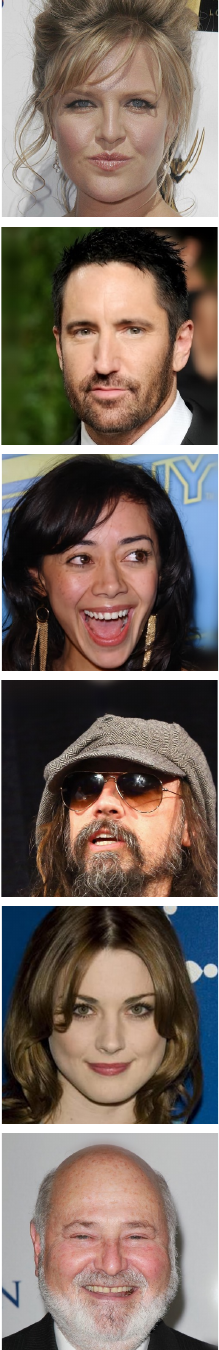 } 
\subcaption*{HQ}
\end{subfigure}
\hspace{-4mm}
% \vspace{-3mm}
\caption{More qualitative comparison results on \textbf{CelebA-Test}. (Zoom in for best view).}
\label{fig:supp_syn}
\end{figure*}

\begin{figure*}[h]
\begin{subfigure}{0.123\textwidth}
\includegraphics[height=6cm]{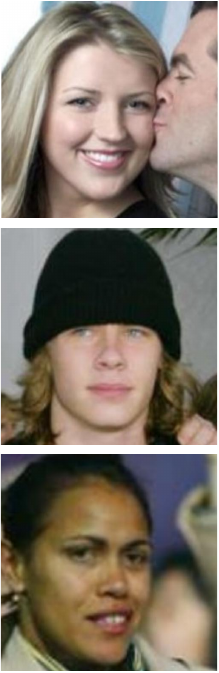} 
\subcaption*{LQ}
\end{subfigure}
\hspace{-4mm}
\begin{subfigure}{0.123\textwidth}
\includegraphics[height=6cm]{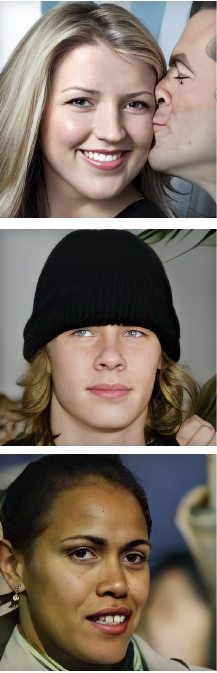} 
\subcaption*{GPEN~\cite{yang2021gpen}}
\end{subfigure}
\hspace{-4mm}
\begin{subfigure}{0.123\textwidth}
\includegraphics[height=6cm]{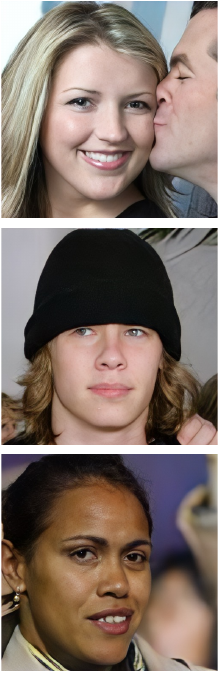} 
\subcaption*{GFPGAN~\cite{wang2021gfpgan}}
\end{subfigure}
\hspace{-4mm}
\begin{subfigure}{0.123\textwidth}
\includegraphics[height=6cm]{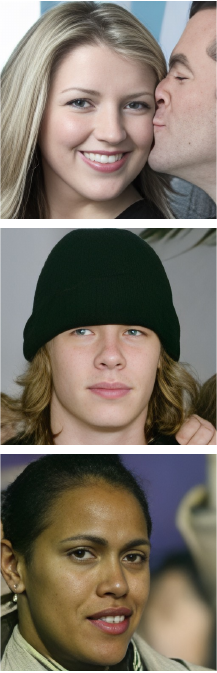} 
\subcaption*{VQFR~\cite{gu2022vqfr}}
\end{subfigure}
\hspace{-4mm}
\begin{subfigure}{0.124\textwidth}
\includegraphics[height=6cm]{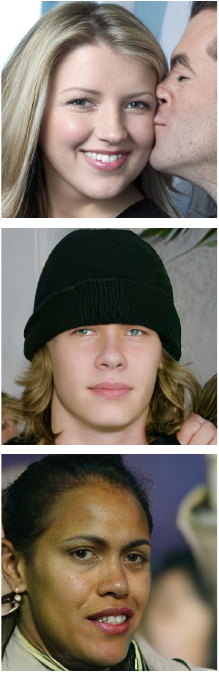} 
\subcaption*{\scriptsize{RestoreFormer}~\cite{wang2022restoreformer}}
\end{subfigure}
\hspace{-4mm}
\begin{subfigure}{0.123\textwidth}
\includegraphics[height=6cm]{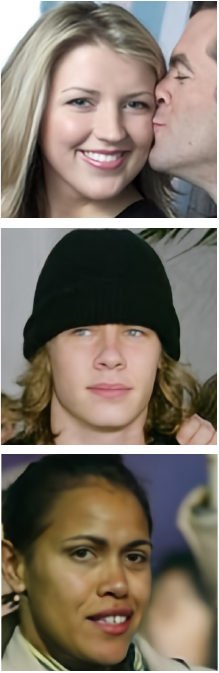} 
\subcaption*{SwinIR~\cite{liang2021swinir}}
\end{subfigure}
\hspace{-4mm}
\begin{subfigure}{0.123\textwidth}
\includegraphics[height=6cm]{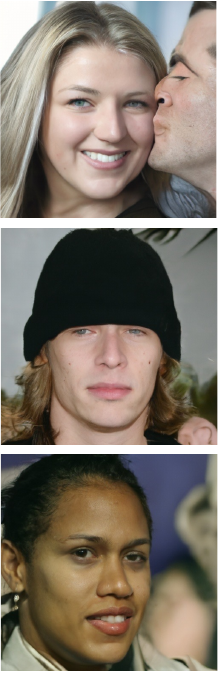} 
\subcaption*{DifFace~\cite{yue2022difface}}
\end{subfigure}
\hspace{-4mm}
\begin{subfigure}{0.123\textwidth}
\includegraphics[height=6cm]{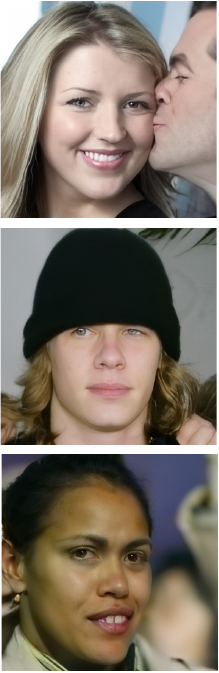} 
\subcaption*{DR2~\cite{wang2023dr2}}
\end{subfigure}
\hspace{-4mm}
\begin{subfigure}{0.123\textwidth}
\includegraphics[height=6cm]{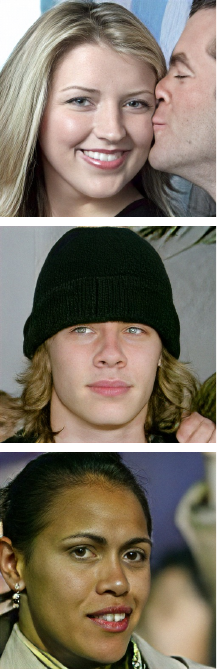} 
\subcaption*{\textbf{Ours}}
\end{subfigure}
\hspace{-4mm}
% \vspace{-2mm}
\caption{More qualitative comparison results on \textbf{LFW-Test}. (Zoom in for best view).}
\label{fig:supp_lfw}
\end{figure*}

\begin{figure*}[]
\begin{subfigure}{0.123\textwidth}
\includegraphics[height=6cm]{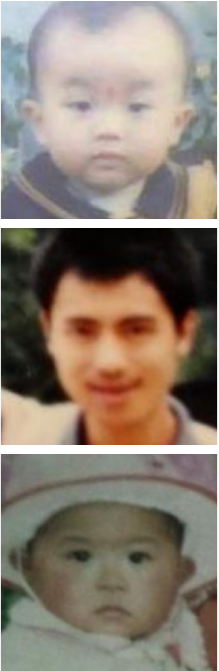} 
\subcaption*{LQ}
\end{subfigure}
\hspace{-4mm}
\begin{subfigure}{0.123\textwidth}
\includegraphics[height=6cm]{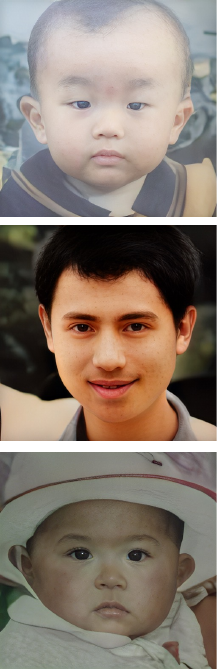} 
\subcaption*{GPEN~\cite{yang2021gpen}}
\end{subfigure}
\hspace{-4mm}
\begin{subfigure}{0.123\textwidth}
\includegraphics[height=6cm]{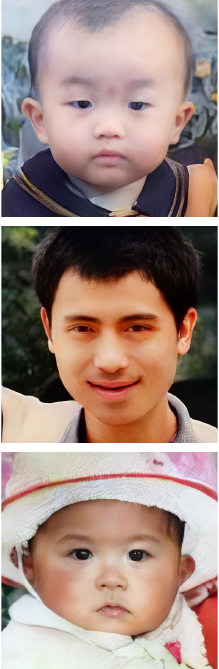} 
\subcaption*{GFPGAN~\cite{wang2021gfpgan}}
\end{subfigure}
\hspace{-4mm}
\begin{subfigure}{0.123\textwidth}
\includegraphics[height=6cm]{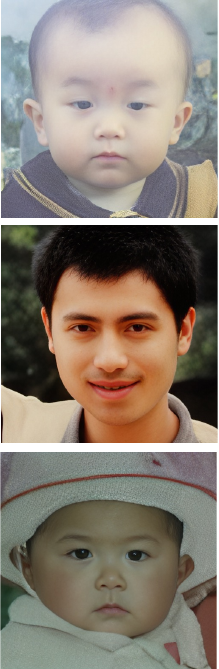} 
\subcaption*{VQFR~\cite{gu2022vqfr}}
\end{subfigure}
\hspace{-4mm}
\begin{subfigure}{0.124\textwidth}
\includegraphics[height=6cm]{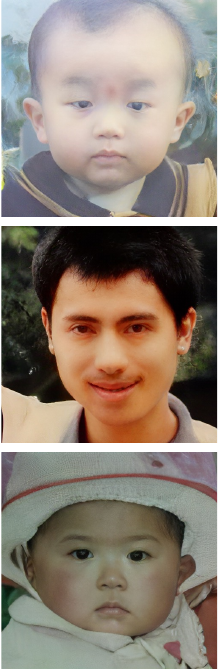} 
\subcaption*{\scriptsize{RestoreFormer}~\cite{wang2022restoreformer}}
\end{subfigure}
\hspace{-4mm}
\begin{subfigure}{0.123\textwidth}
\includegraphics[height=6cm]{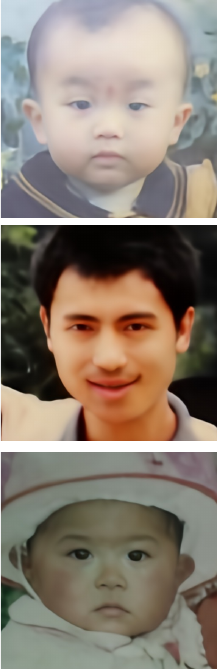} 
\subcaption*{SwinIR~\cite{liang2021swinir}}
\end{subfigure}
\hspace{-4mm}
\begin{subfigure}{0.123\textwidth}
\includegraphics[height=6cm]{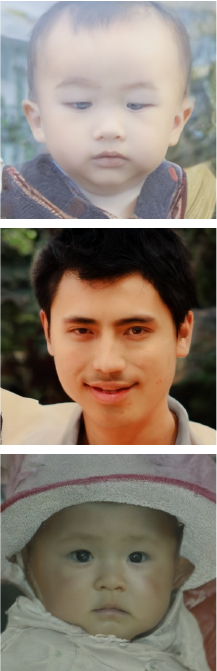} 
\subcaption*{DifFace~\cite{yue2022difface}}
\end{subfigure}
\hspace{-4mm}
\begin{subfigure}{0.123\textwidth}
\includegraphics[height=6cm]{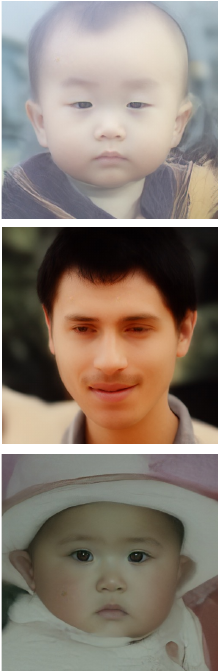} 
\subcaption*{DR2~\cite{wang2023dr2}}
\end{subfigure}
\hspace{-4mm}
\begin{subfigure}{0.123\textwidth}
\includegraphics[height=6cm]{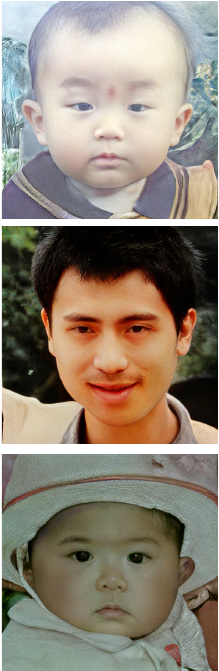} 
\subcaption*{\textbf{Ours}}
\end{subfigure}
\hspace{-4mm}
\vspace{-2mm}
\caption{More qualitative comparison results on \textbf{WebPhoto-Test}. (Zoom in for best view).}
\label{fig:supp_web}
\end{figure*}

\begin{figure*}[]
\begin{subfigure}{0.123\textwidth}
\includegraphics[height=6cm]{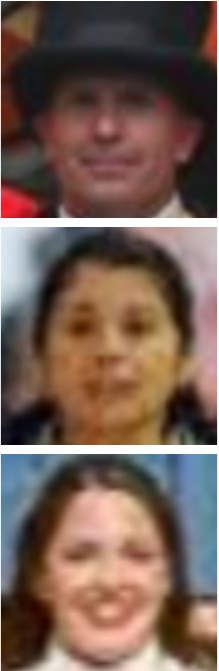} 
\subcaption*{LQ}
\end{subfigure}
\hspace{-4mm}
\begin{subfigure}{0.123\textwidth}
\includegraphics[height=6cm]{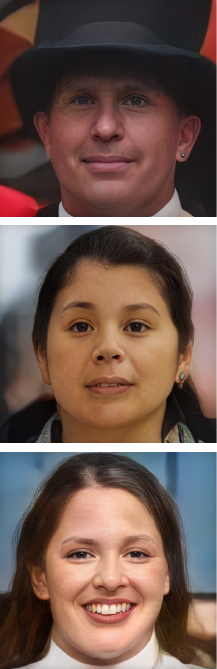} 
\subcaption*{GPEN~\cite{yang2021gpen}}
\end{subfigure}
\hspace{-4mm}
\begin{subfigure}{0.123\textwidth}
\includegraphics[height=6cm]{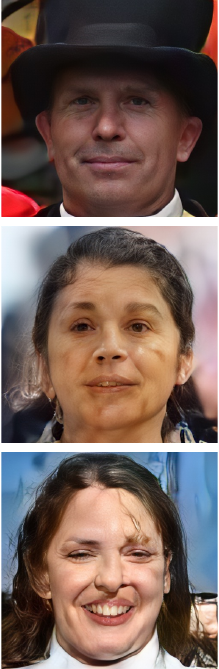} 
\subcaption*{GFPGAN~\cite{wang2021gfpgan}}
\end{subfigure}
\hspace{-4mm}
\begin{subfigure}{0.123\textwidth}
\includegraphics[height=6cm]{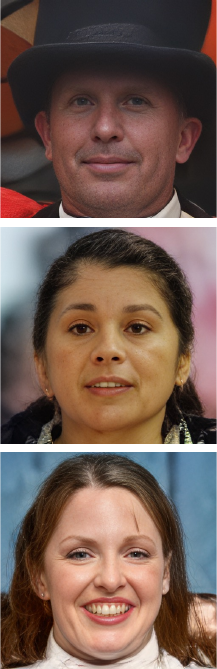} 
\subcaption*{VQFR~\cite{gu2022vqfr}}
\end{subfigure}
\hspace{-4mm}
\begin{subfigure}{0.124\textwidth}
\includegraphics[height=6cm]{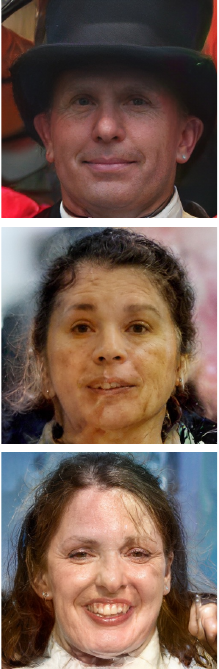} 
\subcaption*{\scriptsize{RestoreFormer}~\cite{wang2022restoreformer}}
\end{subfigure}
\hspace{-4mm}
\begin{subfigure}{0.123\textwidth}
\includegraphics[height=6cm]{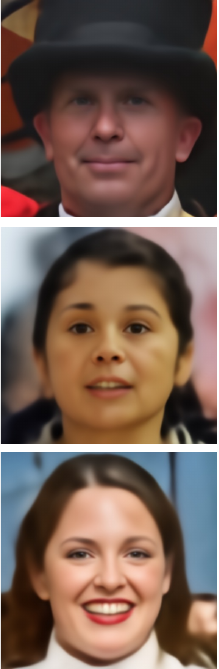} 
\subcaption*{SwinIR~\cite{liang2021swinir}}
\end{subfigure}
\hspace{-4mm}
\begin{subfigure}{0.123\textwidth}
\includegraphics[height=6cm]{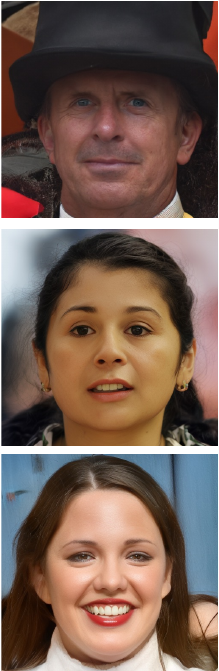} 
\subcaption*{DifFace~\cite{yue2022difface}}
\end{subfigure}
\hspace{-4mm}
\begin{subfigure}{0.123\textwidth}
\includegraphics[height=6cm]{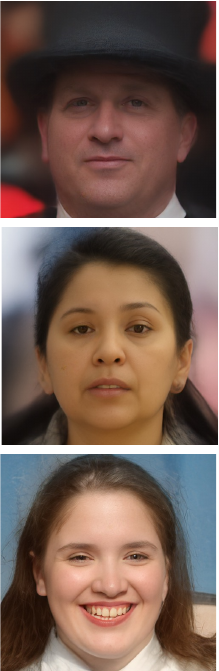} 
\subcaption*{DR2~\cite{wang2023dr2}}
\end{subfigure}
\hspace{-4mm}
\begin{subfigure}{0.123\textwidth}
\includegraphics[height=6cm]{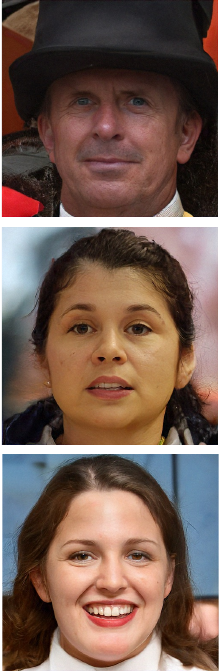} 
\subcaption*{\textbf{Ours}}
\end{subfigure}
\hspace{-4mm}
% \vspace{-2mm}
\caption{More qualitative comparison results on \textbf{WIDER-Test}. (Zoom in for best view).}
\label{fig:supp_wider}
\end{figure*}

More qualitative comparison results are illustrated in \cref{fig:supp_syn} $\sim$ \cref{fig:supp_wider}.
For CelebA-Test (\cref{fig:supp_syn}), with increasing degradation applied (from top to bottom), our method can generate authentic facial images while well preserving the identity.
Qualitative comparison on real-world datasets: LFW-Test (\cref{fig:supp_lfw}), WebPhoto-Test(\cref{fig:supp_web}) and WIDER-Test (\cref{fig:supp_wider}) shows that our method (last column) can restore photorealistic images without destroying style and color of the original image.
Meanwhile, more fine-grained facial details are recovered such as the texture of the beanie (Row 2 in \cref{fig:supp_lfw}), beard (Row 2 in \cref{fig:supp_web}), and wrinkles (Row 1 in \cref{fig:supp_wider}). 

%-------------------------------
%-------------------------------

%-------------------------------
%-------------------------------

%-------------------------------
%-------------------------------

%-------------------------------
%-------------------------------